\documentclass[accepted]{uai2025} 
                        
\usepackage[british]{babel}

\usepackage{natbib} 
\usepackage{mathtools} 
\usepackage{tikz} 

\usepackage{comment}
\usepackage{bm}
\usepackage{enumitem}
\usepackage{amsmath}        
\usepackage{amssymb}
\usepackage{pifont}
\newcommand{\cmark}{\ding{51}}%
\newcommand{\xmark}{\ding{55}}%

\usepackage{wrapfig, lipsum}
\usepackage{adjustbox}
\usepackage{dblfloatfix}

\usepackage{graphicx}
\usepackage{multirow, booktabs}
\usepackage{makecell}
\usepackage{subcaption}
\usepackage{bbm}
\usepackage{booktabs}

\usepackage{scalerel} 
\usepackage{tikz}

\newcommand{\std}[1]{\scriptsize{$\pm$ #1}}

\newcommand{\n}{\textbackslash n}

\DeclareMathOperator*{\argmax}{arg\,max}
\DeclareMathOperator*{\argmin}{arg\,min}

\newcommand{\tP}{{\tt P}}
\newcommand{\tp}{{\tt p}}

\title{Generalised Probabilistic Modelling and Improved Uncertainty Estimation in Comparative LLM-as-a-judge}

\author[1]{\href{mailto:<yf286@cam.ac.uk>?Subject=L-EDD UAI 2023}{Yassir~Fathullah}{}}
\author[1]{Mark~J.~F.~Gales}
\affil[1]{%
    Engineering Department\\
    University of Cambridge\\
    UK
}

\begin{document}
\maketitle

\begin{abstract}
    This paper explores generalised probabilistic modelling and uncertainty estimation in comparative LLM-as-a-judge frameworks. We show that existing Product-of-Experts methods are specific cases of a broader framework, enabling diverse modelling options. Furthermore, we propose improved uncertainty estimates for individual comparisons, enabling more efficient selection and achieving strong performance with fewer evaluations. We also introduce a method for estimating overall ranking uncertainty. Finally, we demonstrate that combining absolute and comparative scoring improves performance.
    Experiments show that the specific expert model has a limited impact on final rankings but our proposed uncertainty estimates, especially the probability of reordering, significantly improve the efficiency of systems reducing the number of needed comparisons by $\sim\!50\%$. Furthermore, ranking-level uncertainty metrics can be used to identify low-performing predictions, where the nature of the probabilistic model has a notable impact on the quality of the overall uncertainty.
\end{abstract}

\section{Introduction}
\label{sec:intro}



Instruction-tuned Large Language Models (LLMs) have shown impressive zero-shot performance on a wide range of natural language processing and generation tasks \citep{ouyang2022training, lima2023zhou, chung2022scaling}.
While the number of downstream applications of aligned LLMs increases \citep{brown2020language, achiam2023gpt, dubey2024llama3}, so does the need to evaluate their performance on bespoke tasks, which could lack labelled data or are costly for humans to judge at scale \citep{zheng2023judging, wang2022self, taori2023stanford}.
As an alternative, instruction-tuned LLMs have increasingly been used as a replacement for humans to evaluate the quality of natural language generations that demonstrate high correlations with human judgements \citep{zheng2023judging, liusie-etal-2024-llm, bubeck2023sparks, wang2023large, chiang2023can}.

There are two standard approaches to using LLM in judging responses.
\textbf{Absolute scoring}: Prompt an LLM to evaluate a certain response attribute on a defined scale (e.g., 1 to 10). \textbf{Comparative scoring}: Prompt an LLM to choose which of the two responses to a given query displays higher quality for a given attribute.
Absolute scoring is a straightforward and effective method for evaluating a variety of responses to a query and ranking them. However, the scores obtained are unreliable and may vary significantly between different LLM judges. 
Alternatively, the more expensive comparative scoring approach has consistently demonstrated higher correlations with human judgements \citep{zheng2023judging, liusie-etal-2024-llm, qin2023large}. However, a drawback of this approach is that it scales quadratically with the number of response candidates, which can become prohibitively expensive due to the inference costs of LLMs.

To address the computational limitations of comparative assessment, various approaches have been proposed. Notably, it is possible to extract more information from the LLM-as-a-judge than just a binary decision. Various works have, as opposed to using simple win-ratio, resorted to using the average probability output from the LLM \citep{qin2023large, zheng2023judging, liusie-etal-2024-llm, park2024paireval}. 
Building on this idea, \citet{liusie2024efficient} introduced a Product-of-Experts (PoE) \citep{hinton1999products, Welling:2007} framework in modelling comparative scoring. In principle, the joint distribution of candidate scores can be broken down into arbitrarily chosen experts that model the score differences of two instances at a time, allowing for a partial set of comparisons to model the full joint distribution. This directly allows one to obtain a ranking of candidates without having to perform all possible comparisons, showing that only a fraction of the total number of comparisons is needed to obtain highly competitive performance.

\textbf{Contributions:} In this paper, we generalise the expert in the comparative framework and derive how there is a wide range of viable options. Starting from the Beta distribution in modelling LLM probabilities, we show that prior works are specific instances of this choice. 
We also propose improved estimates for the uncertainty in individual comparisons and show how these updated uncertainties allow us to make even fewer comparisons without loss of performance.
In addition to the uncertainty of a comparison, we also propose an uncertainty in the overall ranking of a set of candidates.
Finally, we show that the Product-of-Experts framework easily lends itself to combinations of various scoring approaches. Specifically, we show that absolute scoring can complement comparative scoring efficiently and cheaply and lead to improved overall performance.

\section{Background and Related Work}


\textbf{NLG Evaluation with LLMs:} The extensive natural language generation (NLG) capabilities of instruction-tuned large language models \citep{achiam2023gpt, ouyang2022training, chung2022scaling, dubey2024llama3} have prompted recent work on open-ended generation evaluation using LLMs. Methods such as GPTScore \citep{fu2023gptscore} rank responses based on the likelihood of generation and G-Eval \citep{liu-etal-2023-g} which uses chain-of-thought and form-filling to evaluate the quality of a response. Furthermore, LLM-as-a-judge \citep{zheng2023judging} approaches score responses on an absolute scale \citep{wang2023chatgpt, kocmi2023large} or comparative manner by comparing responses against each other \citep{qin2023large, liusie-etal-2024-llm, liusie2024efficient} and building an overall ranking through the set of pairwise comparisons.

\textbf{LLM-Based Comparative Assessment:} The work by \citet{liusie-etal-2024-llm} showed that comparative assessment yields superior performance compared to absolute scoring methods and various custom baselines. By making all possible $N(N\!-\!1)$ pairwise comparisons of $N$ candidate responses, and computing the win-ratio, an overall ranking can be obtained. This style of approach has found its applications in many places. \citet{qin2023large} utilised pairwise comparisons to retrieve relevant sources, using both the full set of comparisons and sorting-based algorithms. \citet{park2024paireval} employed comparative assessment for dialogue evaluation, calculating the average probability across a randomly sampled set of comparisons to determine score quality. Finally, \citet{liu2024aligning} demonstrated the limitations of LLM scoring, and resorted to using pairwise comparisons. They introduced PAirwise-preference Search, a variation of the merge sort algorithm which utilises LLM probabilities.

\textbf{Ranking from Pairwise Comparisons:} The idea of generating a full ranking from pairwise comparisons has been extensively studied. Arguably the most well-known example of this is the ranking of tennis players based on the outcomes of games and this kind of problem has applications in many different areas. 
Anything from sports and gaming \citep{beaudoin2018computationally, csato2013ranking}, web search \citep{dwork2001rank}, social studies \citep{manski1977structure, louviere2000stated} to information retrieval \citep{cao2007learning, liu2009learning} requires modelling through pairwise events. 
The most common approach to model pairwise comparisons is the Bradley-Terry (BT) model \citep{bradley1952rank}. By assigning each candidate a latent score, the probability that one candidate wins over another is based on the underlying score difference. The latent scores can then be obtained by maximising the log-likelihood of the data \citep{david1963method, cattelan2012models}. 
Finally, the TrueSkill model \citep{herbrich2006trueskill, minka2018trueskill} generalises the Bradley-Terry model by incorporating uncertainties in candidate scores within a Bayesian framework in a sports context.

\textbf{Product-of-Experts:} Each comparison $\mathcal{C}_{k} = (i, j, p_{ij})$ contains the ids of the candidates being compared and the corresponding probability produced by the LLM that it think $i$ is better than $j$ for a certain attribute. The comparison $\mathcal{C}_k$ between a pair of candidates then provides information about the distribution of scores $\bm s = s_{1:N}$ of the candidates. 

The Product-of-Experts (PoE) \citep{hinton1999products} approach presents a simple and effective way to combine the information from multiple comparisons according to:
\begin{equation*}
    \tp(\bm s \vert \mathcal{C}_{1:K}) \propto \prod_k \tp(\bm s \vert \mathcal{C}_k) = \prod_k \tp(s_i - s_j \vert \mathcal{C}_k)
\end{equation*}
The distribution can be simplified into a product of individual experts, and be further simplified as it involves comparisons made exclusively between pairs of candidates. 
The work of \citet{liusie2024efficient} proposed the soft Bradley-Terry expert as an extension to the standard BT model:
\begin{equation*}
    \tp(s_i - s_j \vert \mathcal{C}_k) \hspace{-0.5mm}\propto\hspace{-0.5mm} \sigma(s_i-s_j)^{p_{ij}} (1 - \sigma(s_i-s_j))^{1 - p_{ij}}
\end{equation*}
where the probability $p_{ij}$ is obtained from the LLM when comparing candidate $i$ against $j$. The more experts/comparisons are included, the better the resulting estimate of the scores should be.
Scores can then be retrieved by optimising $\tp(\bm s \vert \mathcal{C}_{1:K})$ using iterative \citep{zermelo1929berechnung, newman2023efficient} or standard gradient-based approaches.

\section{Generalised Expert Modelling}
\label{sec:poe}

This section will focus on the nature of the expert for modelling both absolute and pairwise comparisons. We generally have $N$ candidates with associated scores $s_{1:N}$. Given a partial set of comparisons $\mathcal{C}_{1:K}$, the aim is to predict a set of scores $\hat{s}_{1:N}$ which ranks the candidates as closely as possible to the true ranking.
The following section will then focus on how to find an optimal set of comparisons $\mathcal{C}_{1:K}$ through uncertainty estimation, aiming to achieve better correlation with the true scores using fewer comparisons.

\subsection{Comparative Expert Modelling}
\label{ssec:comparative}

The experts in prior work all modelled the score difference of candidates $\tp\left(s_i - s_j \vert \mathcal{C}_{k}\right)$.
Through a simple change of variables we propose a more generalised version of the expert in comparative modelling:
\begin{align*}
    \tp\left(s_i - s_j \vert \mathcal{C}_{k}\right) = f'(s_i - s_j) \tilde\tp\big(f(s_i - s_j) \big\vert \mathcal{C}_{k}\big)
\end{align*}
where the $f(\cdot)$ is a generic monotonically increasing function. Any choice of $f$ is viable as long as the distribution $\tilde{\tp}$ supports it. By letting $f$ be the sigmoid function and using an underlying Beta distribution $\tilde\tp\big(f \big\vert \mathcal{C}_{k}\big) = \mathcal{B}\big(f; p_{ij}, 1 - p_{ij}\big)$ we regain the soft Bradley-Terry model from \cite{liusie2024efficient}. However, from this point, it is clear that there are many more viable options for modelling a pairwise event. In this work, we will investigate several other combinations starting with a sigmoid and a general:
\begin{align}
    \label{eq:gen beta}
    \tilde\tp\big(f \big\vert \mathcal{C}_{k}\big) = \mathcal{B}(f; p_{ij} + \alpha, 1 - p_{ij} + \beta)
\end{align}
model. We will also try an unconventional choice by combining the Gaussian distribution with a sigmoid-like function:
\begin{align}
    \nonumber
    \tp\left(s_i - s_j \vert \mathcal{C}_{k}\right) = \hspace{40mm}\\
    \label{eq:gen gauss}
    \mathcal{N}(s_i-s_j;0, 1) \mathcal{N}\big(\Phi(s_i-s_j);p_{ij}, 1\big)
\end{align}
where $\Phi(\cdot)$ is the cumulative density of the Gaussian. Ablations will study the impact of the choice of $f$ and the underlying distribution $\tilde\tp$.


\subsection{Expert Combinations}
\label{ssec:combination}

As has been mentioned several times, absolute scoring is cheaper but worse than comparative scoring. However, it is possible that absolute scoring can provide complementary information so we propose combining the two approaches into a single model:
\begin{equation*}
    \tp(\bm s \vert \mathcal{C}_{1:K}, \mathcal{A}_{1:N}) \hspace{-0.5mm}\propto\hspace{-0.5mm} \prod_k \tp(s_i - s_j \vert \mathcal{C}_k) \prod_n \tp(s_n \vert \mathcal{A}_n) 
\end{equation*}
where $\mathcal{A}_n$ contains the information from an absolute scoring prompt. When prompting an LLM on a scale of 1 to 10 it contains the probabilities of those values.
Unfortunately, absolute scoring provides a discrete score and while it is possible to obtain the associated logits for each value (of 1 to 10) from the LLM (and construct a categorical distribution) it remains difficult to combine the continuous pairwise experts with the discrete absolute experts. Therefore, we also propose using moment matching to transform the categorical expert into a Gaussian $\mathcal{N}(\mu, \sigma^2)$:
\begin{equation*}
    \mu = \sum_{c} cp_c, \hspace{2mm} \sigma^2 = \sum_{c} (c-\mu)^2 p_c
\end{equation*}
where $c$ and $p_c$ represent the class and the associated probability that the LLM would output that class. In our running example, we would have $c \in \{1, \dots, 10\}$. Finally, the absolute expert can be written as:
\begin{align*}
    \tp(s_n \vert \mathcal{A}_n) 
    \approx 
    \mathcal{N}(s_n; \mu_n, \sigma^2_n), \hspace{1mm}\forall n = 1, \dots, N
\end{align*}
and would allow us to operate with continuous values and optimise the skill scores using gradient-based approaches. In this work, we rely solely on simple absolute scoring but there exist many more sophisticated pointwise scoring approaches like G-Eval \citep{liu-etal-2023-g} which can provide further improvements.

\subsection{Home Advantage}
\label{ssec:home}

A big issue plaguing LLM-based approaches is bias in the system. In our case, the probability outputs of an LLM are inconsistent $p_{ij} \neq 1 - p_{ji}$, meaning that the LLM-based judge can assign conflicting probabilities when comparing $i$ to $j$ as opposed to $j$ to $i$. This stems from positional biases in the system \citep{zheng2023judging, chen2024humans, liusie2024teacher, wang-etal-2023-primacy, zhu2023judgelm, chen2024premiseordermattersreasoning, liu-etal-2024-lost}.
To resolve such an issue, we rely on one of the two approaches. \textbf{Permutation Debiasing:} For each comparison we make two LLM calls for both $i$ vs $j$ and $j$ vs $i$ to obtain a final debiased probability $\tilde{p}_{ij} = \frac{1}{2}(p_{ij}\!+\!(1\!-\!p_{ji}))$ which would ensure consistency: $\tilde{p}_{ij} = (1-\tilde{p}_{ij})$. \textbf{Home Advantage:} An alternative approach is to directly incorporate the positional bias into the comparative expert model. Since we already know that a certain position will be preferred over another we can introduce a 'home advantage' \citep{agresti1990categorical, caron2012efficient} parameter to model the inconsistency through our function $f$:
\begin{align*}
    f(s_i-s_j; \Delta) = f(s_i - s_j - \Delta)
\end{align*}
While prior approaches have developed the theory for home advantage in specific use-cases such as Bradley-Terry \citep{caron2012efficient} and Gaussian experts \citep{liusie2024efficient}, our parameterisation through the generic function $f$ allows us to straightforwardly incorporate home advantage into any type of expert. Furthermore, while the work of \citet{liusie2024efficient} estimated the advantage parameter $\Delta$ through bespoke rules for each expert, we estimate it by maximising the likelihood $\tp(\bm s \vert \mathcal{C}_{1:K}, \Delta)$.

\section{Uncertainty Estimation}
\label{sec:uncertainty}

This section will explore how to estimate uncertainty when ranking examples. Two levels of uncertainty will be explored.
\textbf{Pairwise uncertainty:} The uncertainty in the score difference of a pair of candidates. By being able to identify the most uncertain pairs, we can understand what comparisons we should perform to improve the performance.
\textbf{Ranking uncertainty:} The uncertainty in the overall ranking of a set of candidates, which can help us in understanding whether the predicted ranking is trustworthy or not.

\subsection{Laplace's approximation}
\label{ssec:laplace}

Due to the complex nature of the Product-of-Experts (PoE) distribution, $\tp(\bm s \vert \mathcal{C}_{1:K})$, deriving analytical expressions for uncertainties is generally intractable. Therefore, following practice in Bayesian inference with complex models, we rely on Laplace's approximation to estimate the posterior distribution and its associated uncertainties. The approach approximates the distribution $\tp(\bm s \vert \mathcal{C}_{1:K})$ with a Gaussian:
\begin{align*}
    \tp(\bm s \vert \mathcal{C}_{1:K}) \approx \mathcal{N}\left(\bm s; \bm\mu^{(K)}, \bm\Sigma^{(K)}\right)
\end{align*}
where we set (and dropped the superscript for conciseness):
\begin{align*}
    \bm\mu &= \argmax_{\bm s} \hspace{1mm} \ln\tp(\bm s \vert \mathcal{C}_{1:K}) \\
    \bm\Sigma^{-1} &= -\nabla\nabla\hspace{0.5mm}\ln\tp(\bm s \vert \mathcal{C}_{1:K})\big\vert_{\bm \mu}
\end{align*}
There are more advanced approaches to approximating an intractable distribution but we will rely on this simple and efficient scheme in this work. 

\subsection{Pairwise Uncertainty Estimation}
\label{ssec:pairwise uncertainty}

Being able to estimate the uncertainty in a pair of candidates, in their score difference, can allow us to decide which comparisons are useful and which are not. The better the quality of the uncertainty estimate, the fewer comparisons are needed to achieve good performance.
Following \citet{liusie2024efficient}, the aim is to iteratively add additional comparisons to improve the product-of-experts model and the overall ranking of scores. Starting with a unit Gaussian expert/prior on every score we:
\begin{enumerate}
    \item \textbf{Estimate Uncertainty}: Compute the covariance matrix $\bm\Sigma$ based on the currently selected set of comparisons $\mathcal{C}_{1:K}$ using Laplace approximation.
    \item \textbf{Select Next Comparison}: Use an uncertainty metric and the current covariance matrix $\bm\Sigma$ to select the most informative comparison pair $(i, j)$. In large-scale tasks one can select a batch of comparisons.
    \item \textbf{Perform Comparison}: Obtain the LLM probability for the selected pair $(i, j)$ and add this new comparison to the set $\mathcal{C}_{1:K+1}$.
    \item \textbf{Update Model}: Update the Product-of-Experts model to incorporate the new comparison, which implicitly updates the mean $\bm\mu$ and covariance matrix $\bm\Sigma$ for the next iteration.
\end{enumerate}
Prior work poised that the next comparison that should be selected should induce the \textbf{minimum overall uncertainty} in the resulting distribution, giving the following selection criteria under a soft BT model:
\begin{align}
    \label{eq:min unc}
    \argmax_{i, j} \hspace{1mm} \sigma\!\left(\mu_i\!-\!\mu_j\right) \sigma\!\left(\mu_j\!-\!\mu_i\right)
    \big( \Sigma_{ii} - 2\Sigma_{ij} + \Sigma_{jj} \big)
\end{align}
We propose and evaluate two additional pairwise uncertainty metrics which apply to any expert modelling choice. 
\textbf{Variance in Score Difference}: A straightforward measure of uncertainty in a pair of candidates $i$ and $j$ is the variance of their score difference under the approximated Gaussian posterior. Given the covariance matrix $\bm\Sigma$, the variance of the score difference $s_i - s_j$ can be directly computed as $\mathbb{V}[s_i - s_j\vert\mathcal{C}_{1:K}] = \Sigma_{ii} - 2\Sigma_{ij} + \Sigma_{jj}$. To select the next comparison, we aim to maximise this variance, choosing the pair $(i, j)$ for which the model is most uncertain about their score difference:
\begin{align}
    \label{eq:var}
    \argmax_{i, j} \hspace{1mm}  \Sigma_{ii} - 2\Sigma_{ij} + \Sigma_{jj}
\end{align}
\textbf{Probability of Reordering}: While variance captures the general uncertainty in score difference, a potentially more relevant metric for ranking is the probability that the ranking between two candidates is reversed. Assuming that the current model predicts candidate $i$ to be better than candidate $j$ (i.e., $\mu_i > \mu_j$), the probability of reordering, i.e., the probability that candidate $j$ is actually better than candidate $i$ given the current comparisons $\mathcal{C}_{1:K}$, can be calculated using the Gaussian cumulative density:
\begin{align*}
    \tP(s_i < s_j\vert\mathcal{C}_{1:K}) = \Phi\left( \frac{\mu_j - \mu_i}{\sqrt{\Sigma_{ii} - 2\Sigma_{ij} + \Sigma_{jj}}}  \right)
\end{align*}
This metric directly quantifies the likelihood of a rank inversion between the pair. As shown in the Appendix \ref{app:reordering}, this can be simplified to a similar selection form since we want to pick examples with the highest reordering probability:
\begin{align}
    \label{eq:prob reord}
    \argmax_{i, j} \hspace{1mm}  \frac{\Sigma_{ii} - 2\Sigma_{ij} + \Sigma_{jj}}{(\mu_i - \mu_j)^2}
\end{align}
This probability of reordering metric is intuitively appealing as it directly targets the goal of improving ranking accuracy by focusing on comparisons that are most likely to correct potential ranking errors. Furthermore, it naturally encourages the exploration of comparisons between candidates with similar skill levels (high-density regions of the score distribution), as these are the pairs where rank inversions are most probable. Furthermore, since all selection mechanisms take the following form:
\begin{align*}
    \argmax_{i, j} \hspace{1mm}  w(\mu_i - \mu_j)\left(\Sigma_{ii} - 2\Sigma_{ij} + \Sigma_{jj}\right)
\end{align*}
we ablate the weight function $w(\cdot)$.

\subsection{Ranking Uncertainty Estimation}
\label{ssec:ranking uncertainty}

In addition to pairwise uncertainties, it is also valuable to assess the overall uncertainty in the predicted ranking of all candidates. While a full probabilistic ranking distribution is complex to compute, we can use the entropy of the approximated Gaussian score distribution as a proxy for ranking uncertainty. The entropy $\mathcal{H}\!\left[\bm s\vert\mathcal{C}_{1:K}\right]$ of a multivariate Gaussian distribution $\mathcal{N}(\bm s; \bm\mu, \bm\Sigma)$ is given by: $N(1 + \ln(2\pi))/2 + \ln(\det(\bm\Sigma))/2$. Lower entropy indicates a more concentrated and certain score distribution, suggesting a more reliable overall ranking. We will use this entropy metric to assess the overall uncertainty of the rankings produced by various models.

\section{Experimental Setup}

\subsection{Datasets}
\label{ssec:dataset}

We mainly perform experiments on the summary evaluation SummEval dataset \cite{fabbri2021summeval} which contains 100 articles, each with 16 machine-generated summaries evaluated on four different attributes: coherency (COH), consistency (CON), fluency (FLU), and relevancy (REL). We will also use the much larger HANNA dataset \cite{chhun2022human} which has 1056 machine-generated stories annotated by humans on six different attributes. These are averaged to a single overall quality score.

\subsection{Methodology}
\label{ssec:methodology}

We will be relying on Flan-T5 \cite{chung2022scaling} and Qwen2.5-Instruct \cite{qwen2.5} systems to evaluate performance on the SummEval and the larger HANNA datasets. Appendix \ref{app:prompting} will detail our choices, how we structure the prompts and how the probabilities are extracted from each model.

\textbf{Probabilistic Models:} In almost all experiments we will rely on the soft Bradley-Terry extension as our baseline expert model. This will be compared against our proposed extensions to this approach: 
(1) The generalised Beta distribution in Eq. (\ref{eq:gen beta}), (2) the extended Gaussian distribution in Eq. (\ref{eq:gen gauss}) and (3) the combination of comparative and absolute outputs in a single PoE model.
We will not include simple baselines such as average win-ratio and average probability since these have been shown to be inferior on a wide range of tasks \citep{liusie2024efficient, raina2024finetuningllmscomparativeassessment}.

\textbf{Iterative Selection:} We will also compare the probabilistic models in an active learning framework where each model needs to select the comparisons that will induce the best performance. The baseline will be the \textbf{minimum uncertainty} approach given in Eq. (\ref{eq:min unc}) using the soft BT model. We will compare this against our proposed \textbf{variance} (Eq. (\ref{eq:var})) and \textbf{probability of reordering} (Eq. (\ref{eq:prob reord})) selection mechanisms.

\textbf{Ranking Uncertainties:} The entropy of Product-of-Experts models will be investigated on how well they correlate with the actual performance of the predicted rankings, and whether they can identify high-performing predictions.

\textbf{Evaluation Metrics:} Since we are interested in predicting the ranking of candidate responses given a context/query, the main performance metric is Spearman rank correlation between the predicted and the human labelled scores. In SummEval ($N = 16$) we perform absolute and comparative scoring and evaluate the average Spearman across all contexts. For HANNA ($N = 1056$) we rank all generated stories. 
Furthermore, we assess the quality of the various comparison-level uncertainties (iterative selection schemes) by the number of comparisons needed to achieve good Spearman rank correlation. This will be defined as the number of comparisons $K$ required to reach within 90\% performance when using the full set of $N(N-1)$ comparisons.
Finally, the ranking-level uncertainties are evaluated using the area under the receiver operating characteristic curve (AUROC) to detect well-performing rankings.

\begin{table*}[h!]
    \centering
    \caption{Spearman rank correlations (\%) and AUROC (\%) for SummEval.}
    \vspace{-3mm}
    \small
    \begin{tabular}{c|cc|ccccc|ccccc}
        \toprule
        \multirow{2}{*}{LLM} & 
        Function & Distribution & 
        \multicolumn{5}{c|}{Spearman Rank (\%)} & \multicolumn{5}{c}{AUROC (\%)} \\
        \multirow{9}{*}{\vspace{-3mm}Flan-T5}
        & $f$ & $\tilde\tp$ & {\tt COH} & {\tt CON} & {\tt FLU} & {\tt REL} & {\textbf{Avg}} & {\tt COH} & {\tt CON} & {\tt FLU} & {\tt REL} & {\textbf{Avg}} \\
        \midrule
        \multirow{9}{*}{\vspace{1mm}(3B)} & $x$ & Gaussian 
        & 49.1 & 45.2 & 32.5 & 42.2 & 42.3
        & 53.4 & 51.8 & 54.9 & 58.4 & 54.6 \\
        & $\sigma$ & Gaussian 
        & 49.2 & 45.3 & 32.5 & 42.2 & 42.3
        & \textbf{65.5} & 67.3 & 60.8 & 65.8 & 64.9 \\
        & $\Phi$ & Gaussian 
        & 49.2 & 45.3 & 32.5 & 42.3 & 42.3
        & 65.0 & 67.3 & 61.0 & 65.9 & 64.8 \\
        \cmidrule(lr){2-3}
        \cmidrule(lr){4-8}
        \cmidrule(lr){9-13}
        & $\sigma$ & Beta 
        & 49.2 & 45.2 & 32.5 & 42.2 & 42.3
        & 63.8 & \textbf{68.3} & 60.6 & 63.8 & 64.1 \\
        & $\Phi$ & Beta 
        & 49.2 & 45.3 & 32.6 & 42.3 & 42.4
        & 64.8 & 67.4 & 60.9 & 65.7 & 64.7 \\
        \cmidrule(lr){2-3}
        \cmidrule(lr){4-8}
        \cmidrule(lr){9-13}
        & \multicolumn{2}{c|}{Comparative ($\sigma$-Beta)} & 
        \multirow{2}{*}{50.7} & 
        \multirow{2}{*}{45.9} & 
        \multirow{2}{*}{32.9} & 
        \multirow{2}{*}{43.4} & 
        \multirow{2}{*}{43.2} & 
        \multirow{2}{*}{64.5} & 
        \multirow{2}{*}{69.9} & 
        \multirow{2}{*}{62.0} & 
        \multirow{2}{*}{67.0} & 
        \multirow{2}{*}{\textbf{65.9}} \\
        & \multicolumn{2}{c|}{+ Absolute Experts} & & & & & \\
        \midrule
        Qwen2.5 (3B) & $\sigma$ & Beta 
        & 48.5 & 48.6 & 40.8 & 46.5 & 46.1
        & 59.8 & 62.8 & \textbf{63.2} & 57.5 & 60.8 \\
        Qwen2.5 (7B) & $\sigma$ & Beta 
        & 50.4 & 48.9 & 38.7 & 50.5 & 47.1
        & 66.7 & 68.2 & 58.5 & \textbf{69.4} & 65.7 \\
        Qwen2.5 (14B) & $\sigma$ & Beta 
        & \textbf{57.4} & \textbf{52.1} & \textbf{47.4} & \textbf{51.1} & \textbf{52.0}
        & 61.7 & 62.0 & 58.4 & 64.8 & 61.7 \\
        \bottomrule
    \end{tabular}
    \vspace{-2mm}
    \label{tab:poe}
\end{table*}

\begin{figure*}[!b]
    \makebox[1.0\textwidth][c]{
    \begin{subfigure}{0.26\textwidth}
        \includegraphics[width=\textwidth]{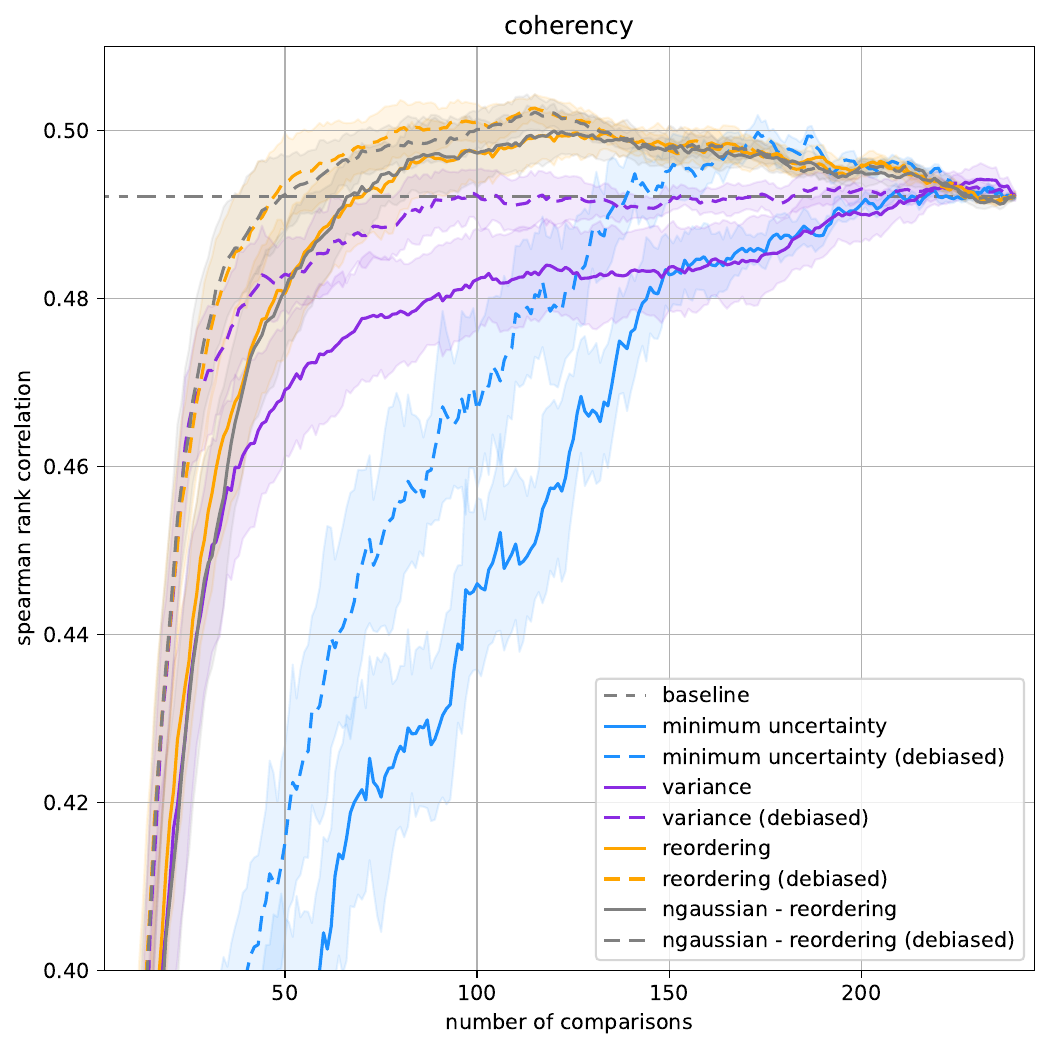}
        \label{fig:sel-coh}
    \end{subfigure}
    \begin{subfigure}{0.26\textwidth}
        \includegraphics[width=\textwidth]{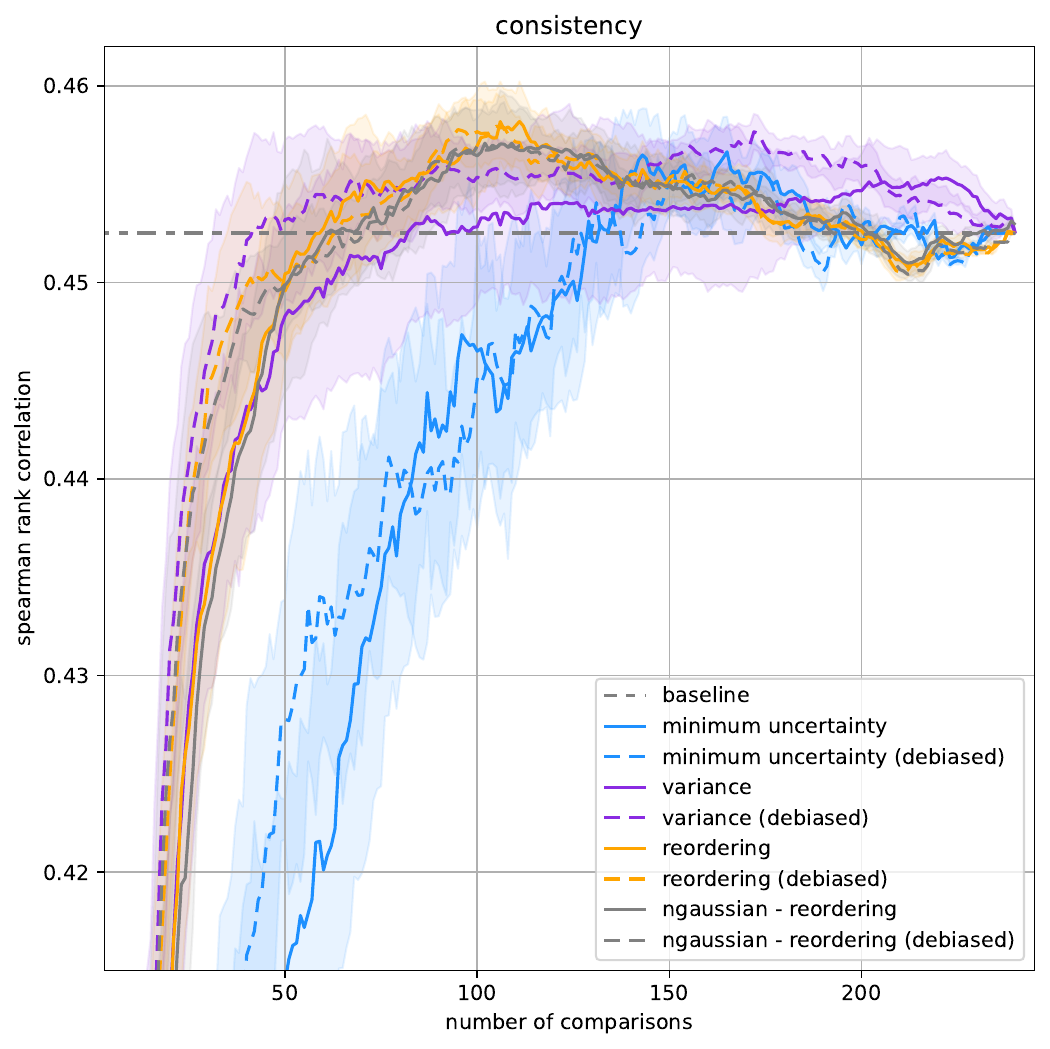}
        \label{fig:sel-con}
    \end{subfigure}
    \begin{subfigure}{0.26\textwidth}
        \includegraphics[width=\textwidth]{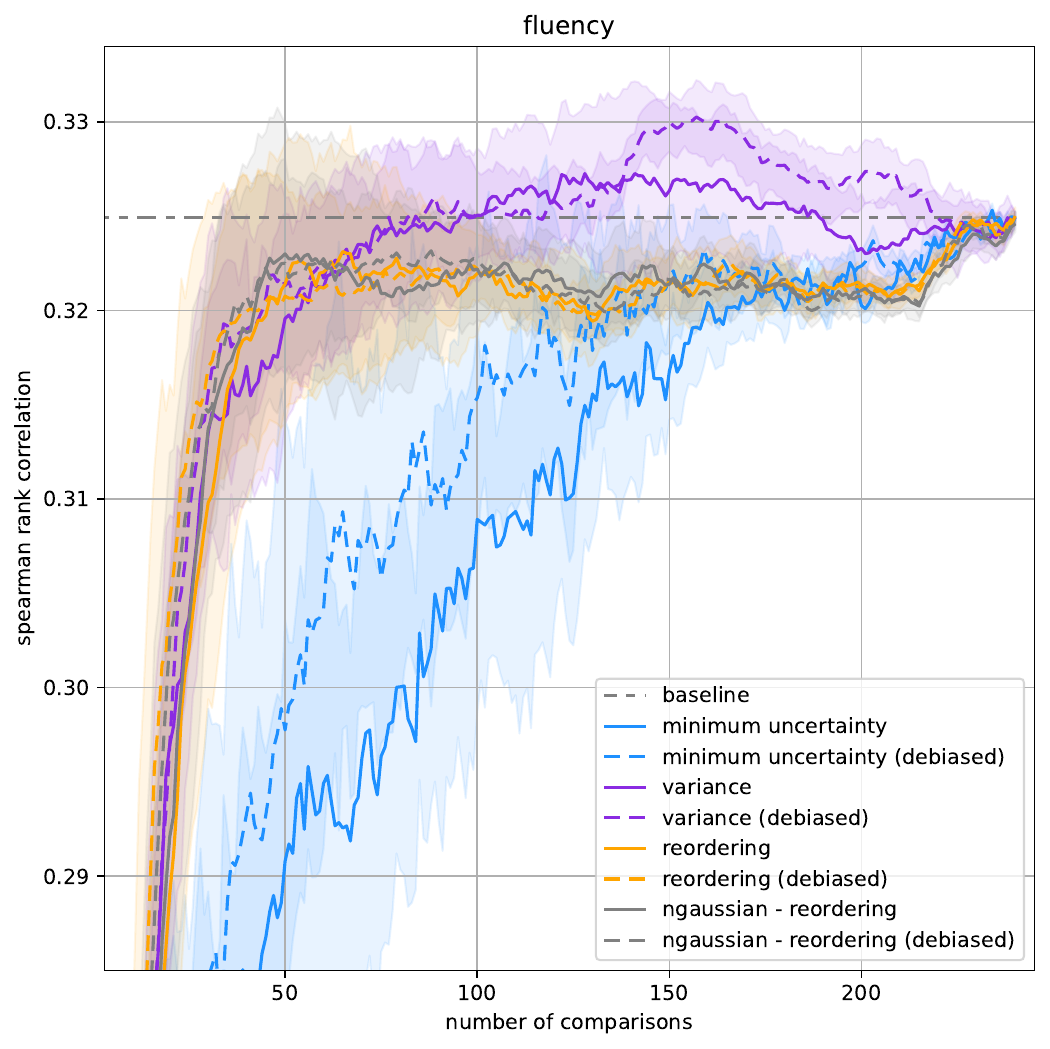}
        \label{fig:sel-flu}
    \end{subfigure}
    \begin{subfigure}{0.26\textwidth}
        \includegraphics[width=\textwidth]{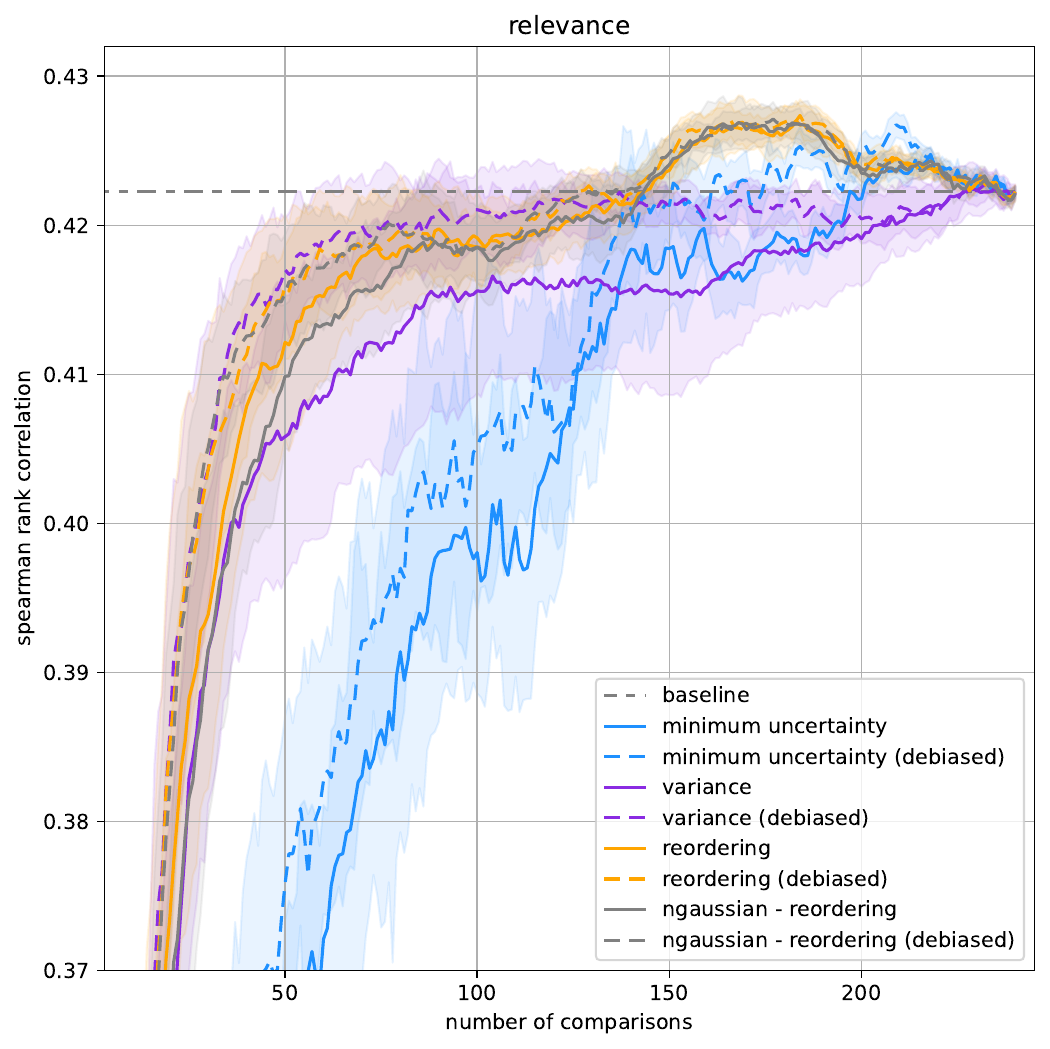}
        \label{fig:sel-rl}
    \end{subfigure}\hspace{3mm}}
    \vspace{-11mm}
    \caption{The Spearman Rank Correlation when iteratively selecting the next examples of lowest confidence/highest uncertainty. The baseline is the soft Bradley-Terry model with the minimum uncertainty metric. We also report the proposed variance and probability of reordering under the soft BT model. Furthermore, the $\Phi$-Gaussian model is referred to as "ngaussian". Debiased refers to permutation debiased probabilities. }
    \label{fig:selection}
\end{figure*}

\section{Results}

\subsection{Form of Distribution}

This section investigates a wide range of PoE models conditioned on the full set of comparisons. For SummEval with $N = 16$ summaries per context, there is a total of $N(N-1) = 240$ number of comparisons. Unless reported otherwise, all results are based on the direct biased outputs of judges.
In the first block of Table \ref{tab:poe}, under a Flan-T5 judge, models showcase similar Spearman performance when evaluated on the full set of comparisons. The only model that performs noticeably better is the combined comparative-absolute model. Absolute scoring seems to extract complementary information and give the overall system a performance boost. Furthermore, improved and larger Qwen LLMs clearly display improved performance.

However, improved performance does not necessarily imply higher quality ranking-level uncertainty estimates. To benchmark, an uncertainty is predicted for each context, meaning to represent how well the predicted ranks perform. The Spearman rank performance of each context is thresholded by the median score and mapped to a binary value so that contexts are classed as '0' or '1'. This allows us to use the standard AUROC score to evaluate detection performance. The second block of Table \ref{tab:poe} shows the AUROC performance of a range of probabilistic models. Unlike previous results, the nature of the probabilistic model seems to have a significant impact on performance. While the performance of various PoE models are similar, the predicted entropy, and by equivalence, the hessian (curvature) of the log-likelihood seems to differ between certain models. This seems to stem mainly from the choice of $f$ but further investigations are needed to understand how to predict robust uncertainties.

The linear-Gaussian model performs significantly worse, only marginally outperforming a random classifier. Furthermore, all models that map skill differences through a sigmoid-like function $f$ perform notably better, with the combined model being the best system similar to previous results. Even the better-performing and larger Qwen systems are unable to outperform the combined model. Furthermore, while Qwen2.5 (3B) displays better Spearman rank than the similarly sized Flan-T5, the uncertainties perform worse. 
While this is only a preliminary investigation into detecting well-performing examples, it is a good start into understanding the problem and what models produce better uncertainties. In Appendix \ref{app:calibration}, we investigate the causes for this and pinpoint some of the problems to an overconfidence issue associated with modern language models.

\begin{table*}[h!]
    \centering
    \caption{\textbf{Measure of efficiency}: The number of comparisons needed to achieve 90\% performance of the baseline soft Bradley-Terry system for each corresponding judge. The baseline is evaluated using all $N(N-1) = 240$ comparisons.}
    \vspace{-3mm}
    \small
    \begin{tabular}{cc|cccc|cccc}
        \toprule
        \multirow{2}{*}{Uncertainties} & \multirow{2}{*}{Debiased} & \multicolumn{4}{c|}{Flan-T5 (3B)} & \multicolumn{4}{c}{Qwen2.5 (3B)} \\
        & & {\tt COH} & {\tt CON} & {\tt FLU} & {\tt REL} & {\tt COH} & {\tt CON} & {\tt FLU} & {\tt REL} \\
        \midrule
        minimum & \xmark 
        & 82.8 \std 9.8 
        & 39.4 \std 6.2 
        & 30.3 \std 9.6 
        & 61.4 \std 8.7 
        & 76.9 \std 8.4 
        & 39.9 \std 3.2 
        & 45.7 \std 6.3 
        & 71.2 \std 7.3 \\
        uncertainty & \cmark 
        & 59.6 \std 8.7
        & 27.1 \std 7.1
        & 23.9 \std 6.1 
        & 46.8 \std 4.6 
        & 57.9 \std 8.3 
        & 35.9 \std 4.2 
        & 43.7 \std 6.8 
        & 63.1 \std 7.9 \\
        \midrule
        \midrule
        \multirow{2}{*}{variance} & \xmark 
        & 29.7 \std 6.9 
        & 17.8 \std 2.7 
        & 13.8 \std 7.6 
        & 27.5 \std 6.7 
        & 44.3 \std 7.1 
        & 24.6 \std 3.6 
        & 27.2 \std 6.6 
        & 53.9 \std 8.0 \\
        & \cmark 
        & 20.3 \std 2.0 
        & 14.3 \std 2.5 
        & 16.7 \std 2.3 
        & 19.7 \std 1.7 
        & 24.4 \std 4.0 
        & 16.7 \std 2.8 
        & 21.8 \std 4.2 
        & 24.5 \std 3.1 \\
        \midrule
        probability & \xmark 
        & 28.1 \std 3.4
        & 17.7 \std 1.9 
        & 18.8 \std 8.1 
        & 22.2 \std 4.2 
        & 29.0 \std 4.5 
        & 23.3 \std 2.2 
        & 24.6 \std 2.4 
        & 34.8 \std 5.6 \\
        of reordering & \cmark 
        & 20.4 \std 1.7 
        & 14.2 \std 2.3 
        & 16.1 \std 3.9 
        & 18.6 \std 2.1 
        & 20.7 \std 1.8 
        & 16.1 \std 2.6 
        & 18.9 \std 3.1 
        & 24.3 \std 4.9 \\
        \bottomrule
    \end{tabular}
    \vspace{-2mm}
    \label{tab:poe-eff}
\end{table*}

\begin{figure*}[!b]
    \makebox[1.0\textwidth][c]{
    \begin{subfigure}{0.34\textwidth}
        \includegraphics[width=\textwidth]{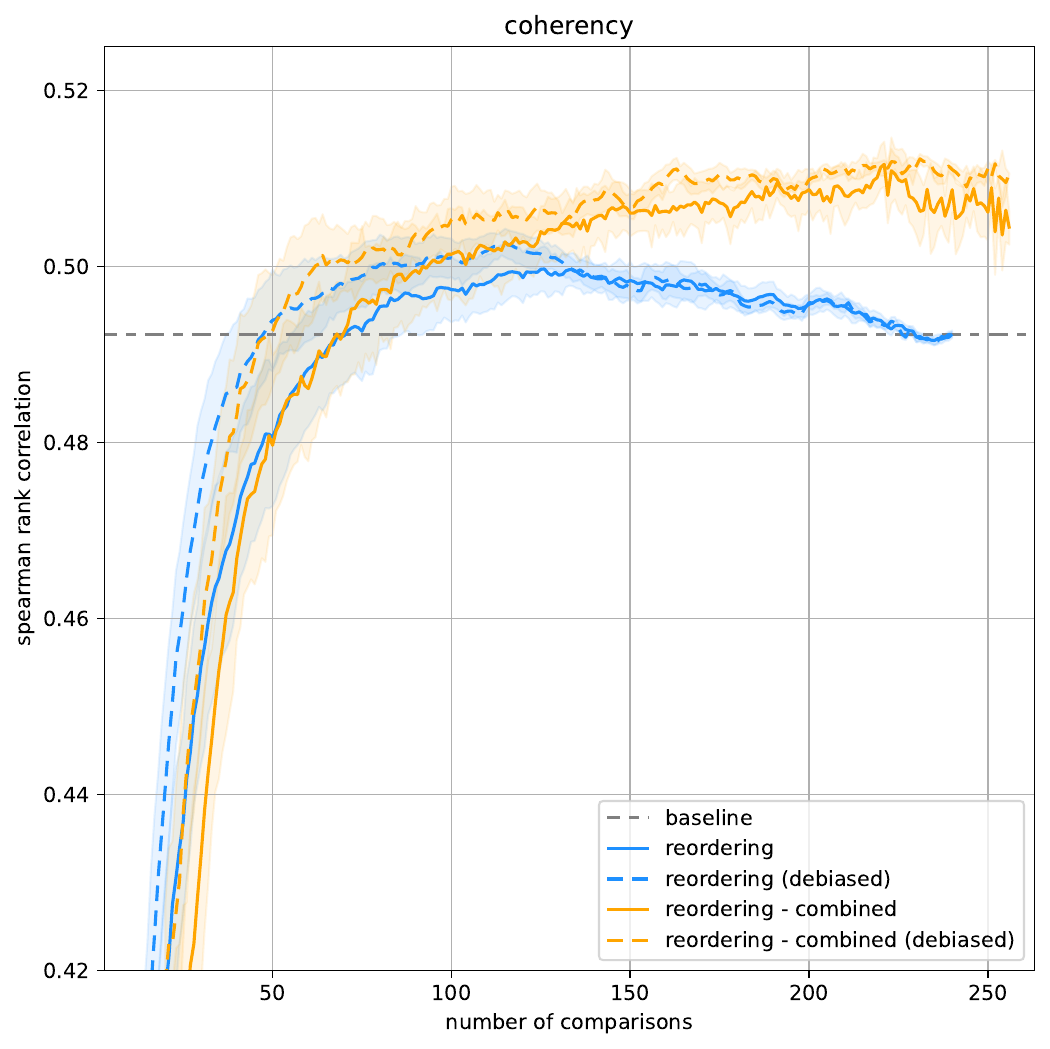}
        \vspace{-6mm}
        \caption{Absolute and comparative combination.}
        \label{fig:selection-combined}
    \end{subfigure}
    \begin{subfigure}{0.34\textwidth}
        \includegraphics[width=\textwidth]{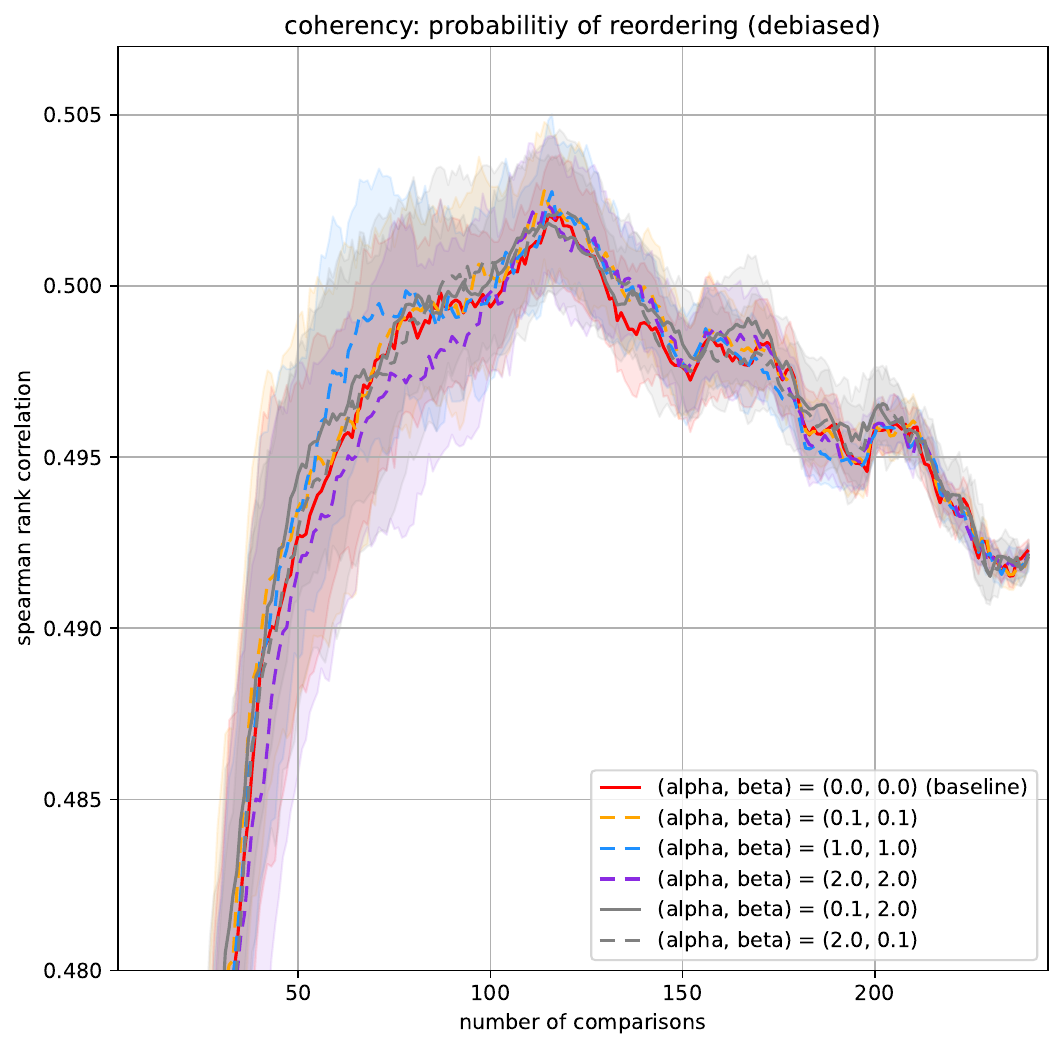}
        \vspace{-6mm}
        \caption{Varying the Beta distribution.}
        \label{fig:selection-beta}
    \end{subfigure}
    \begin{subfigure}{0.34\textwidth}
        \includegraphics[width=\textwidth]{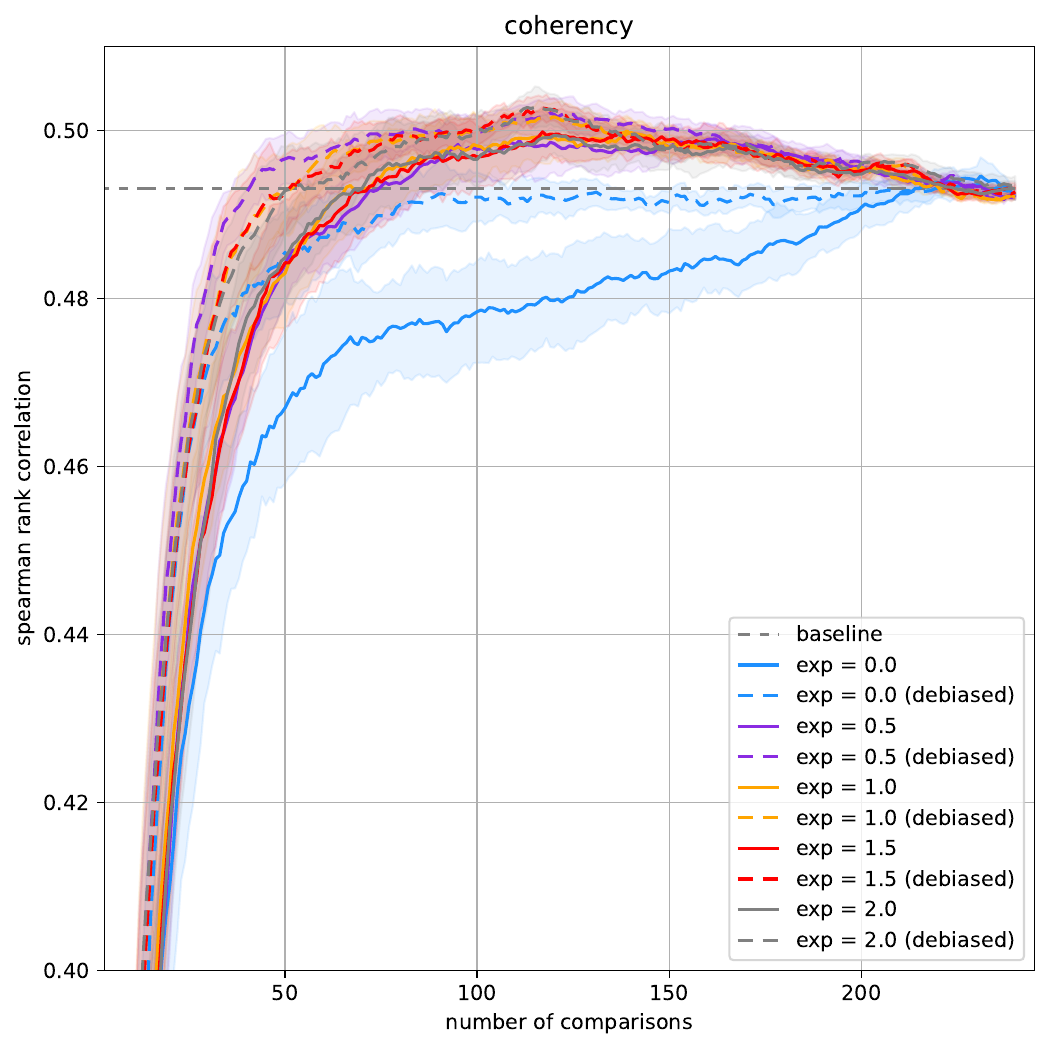}
        \vspace{-6mm}
        \caption{Varying the uncertainty estimate.}
        \label{fig:selection-exp}
    \end{subfigure}\hspace{1mm}}
    \vspace{-7mm}
    \caption{Extended results on Flan-T5: \textbf{(a)} Comparing the soft BT model to the combined model which relies on additional absolute experts. Both use the probability of reordering as the selection criteria. \textbf{(b)} Selection performance. Varying the parameters $\alpha$ and $\beta$ of the Beta distribution in Eq. (\ref{eq:gen beta}). \textbf{(c)} Selection performance. Varying the parameter $\epsilon$ in Eq. (\ref{eq:exp-sweep}).}
    \label{fig:selection-hanna}
\end{figure*}

\subsection{Iterative Selection}
\label{ssec:iterative}

In this section, we explore the quality of various models and uncertainty metrics when iteratively selecting comparisons. Furthermore, this investigation is performed on both biased and debiased probabilities extracted from Flan-T5 (3B). Extended results for Qwen can be found in Appendix \ref{app:results}. Four main points can be observed from the results in Figure \ref{fig:selection}:

\textbf{(1) Model Convergence}: As reported in the table above, all models converge towards the same final performance since the rankings are predicted from the same full set of comparisons.

\textbf{(2) Quality of Uncertainties}: There is a significant gap in performance between various uncertainty metrics. Probability of reordering is shown to outperform the minimum uncertainty metric in all attributes of SummEval for both biased and debiased cases. We expect this performance difference to originate from how each metric was derived. While minimum uncertainty is simply focused on choosing the next comparison that would induce the least uncertainty, the probability of reordering is directly linked to achieving the correct ranking. 

\textbf{(3) Expert Model Invariance}: Probability of reordering under two different models, $\sigma$-Beta (soft BT) and $\Phi$-Gaussian (ngaussian), perform almost identically in all cases. This again reinforces the idea that the nature of the expert model does not matter as much as the uncertainty modelling used in selecting the comparisons. 

\textbf{(4) Performing better with less}: Performance is expected to increase as one adds more and more comparisons to the PoE model. However, in many of the cases above, performance drops until the full set of comparisons is reached. This is related to an overconfidence issue plaguing the uncertainty estimates. We show in Appendix \ref{app:calibration} how the LLM-as-a-judge is miscalibrated, and how temperature annealing is not enough to calibrate and solve the overconfidence issue. 

Furthermore, in Table \ref{tab:poe-eff}, we report the number of examples required to reach 90\% of the final performance when all comparisons have been selected. The baseline 'minimum uncertainty' metric with both biased and debiased probabilities extracted from the LLMs manages to reach the 90\% threshold with at worst 35\% of the number of comparisons. This already represents a significant gain in efficiency, cutting the number of examples to a fraction of the full 240. However, both proposed uncertainty metrics significantly outperform the baseline, reducing the number of required comparisons by an additional $\sim\!60\%$ for Flan-T5 and $\sim\!50\%$ for Qwen. Overall, we observe that the reordering metric is more efficient across datasets and models.

Finally, in Figure \ref{fig:selection-combined}, we compare the soft BT model with the combined comparative-absolute model. Due to the initial cost of obtaining $N = 16$ absolute experts, the combined model initially performs worse. However, both biased and debiased combined models outperform the final soft BT model performance using a fraction of comparisons.

\subsubsection{Ablation Studies}

This section will explore various nuances in the modelling choices. In Figure \ref{fig:selection-beta}, we vary the parameters of the underlying Beta distribution in a generalised soft BT model.

\begin{figure*}[h!]
    \makebox[1.0\textwidth][c]{
    \begin{subfigure}{0.44\textwidth}
        \includegraphics[width=\textwidth]{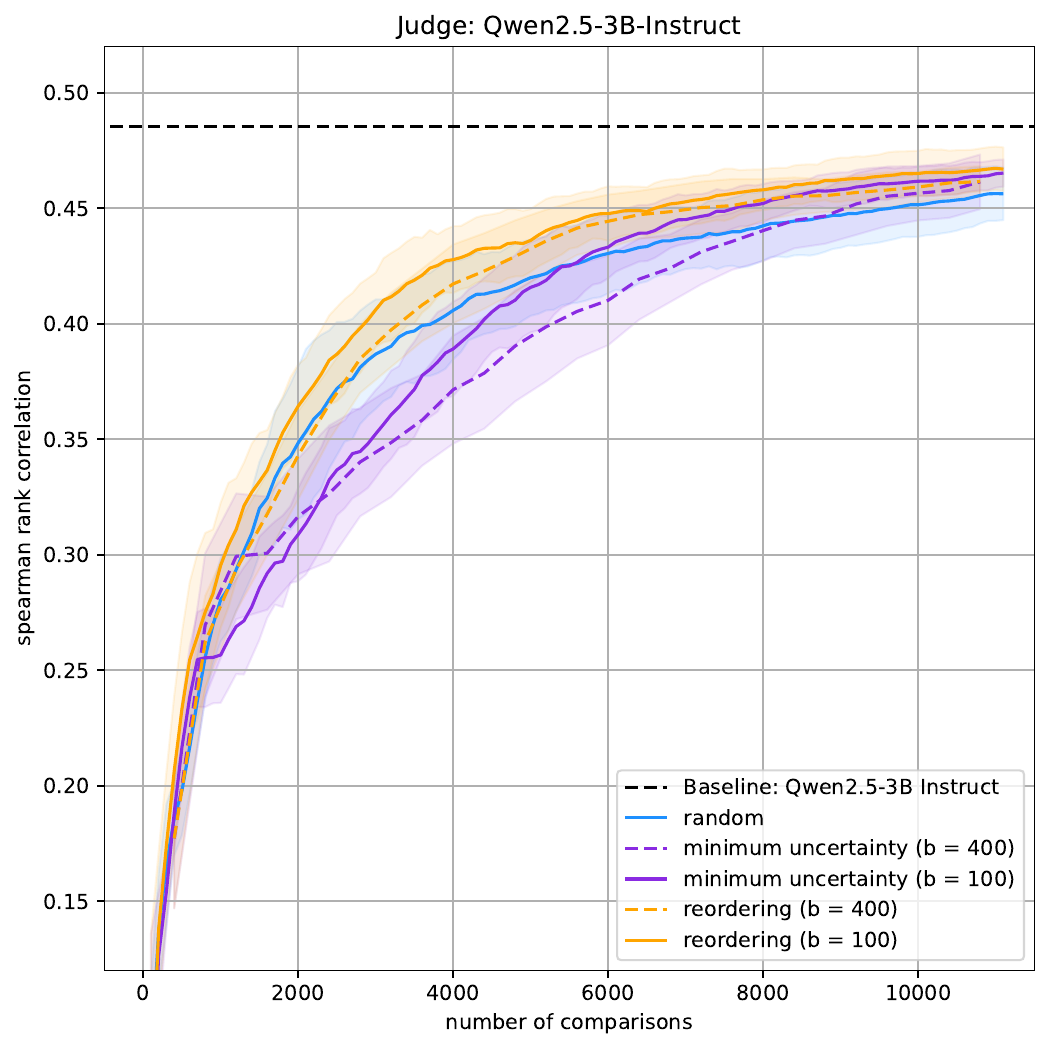}
        \vspace{-6mm}
        \caption{Using Qwen2.5-3B-Instruct as the judge.}
        \label{fig:sel-hanna-3}
    \end{subfigure}
    \hspace{8mm}
    \begin{subfigure}{0.44\textwidth}
        \includegraphics[width=\textwidth]{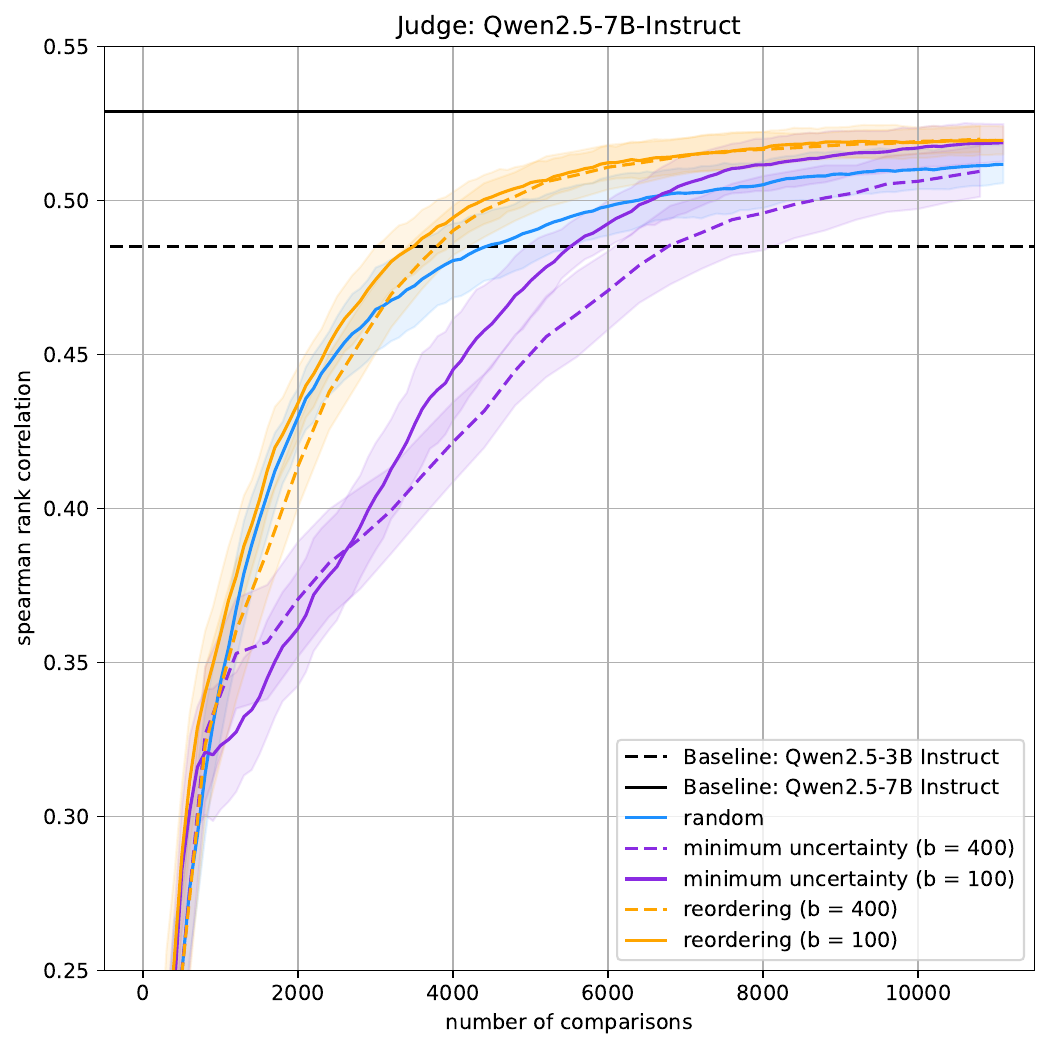}
        \vspace{-6mm}
        \caption{Using Qwen2.5-7B-Instruct as the judge.}
        \label{fig:sel-hanna-7}
    \end{subfigure}}
    \vspace{-7mm}
    \caption{The Spearman Rank Correlation when iteratively selecting the next batch $b$ of examples of lowest confidence/highest uncertainty. The baselines are Qwen2.5-\{3B-7B\} models with all comparisons selected. Furthermore, the efficient baseline is set to the soft BT model with the minimum uncertainty metric.}
    \label{fig:selection-hanna}
\end{figure*}

Evaluated on SummEval ({\tt COH}), it is clear that the underlying Beta distribution has negligible impact on both the selection process and the final performance. Furthermore, we explore generalising the selection metrics with the following:
\begin{align}
    \label{eq:exp-sweep}
    \argmax_{i, j} \frac{\Sigma_{ii} - 2 \Sigma_{ij} + \Sigma_{jj}}{\vert s_i - s_j\vert^\epsilon}
\end{align}
where we vary the exponent $\epsilon$. Setting $\epsilon = 0$ returns variance while $\epsilon = 2$ gives probability of reordering.

The exponent is swept on the same benchmark in Figure \ref{fig:selection-exp}. It is evident that while $\epsilon = 0$ suffers in performance, most of the other values perform similarly. The best-performing option is $\epsilon = 0.5$ which slightly outperforms other options including the probability of reordering.

\subsection{Large-Scale Selection}

All results have been focused on ranking a small set of candidates $N = 16$ which has $K = 240$ possible comparisons. In this section we scale up to the HANNA dataset with $N = 1056$ stories and $K = 1114080$ possible comparisons. Furthermore, due to the cost of iteratively selecting a single comparison, re-estimating Laplace's approximation, and repeating this process, we opt to perform batch acquisitions with $b = \{100, 400\}$. This will showcase in the extreme case how well our best-proposed uncertainty metric, probability of reordering, performs compared to random and minimum uncertainty selection. We will only evaluate the iterative process up until $1\%$ of all possible comparisons to test the efficiency of approaches.

In Figure \ref{fig:selection-hanna}, various iterative schemes are reported. Furthermore, the performance of the soft BT model with all possible comparisons is reported under both backbone judges. In these results, one can observe minimum uncertainty to suffer significantly compared to the simplest baseline of random selection when using both Qwen2.5-3B and 7B as backbone judges. This is caused due to the lack of diversity when selecting a batch of comparisons. While reducing the batch size of acquisitions helps performance, it still lacks significantly compared to the probability of reordering which is far more robust to larger batch acquisitions.

\section{Conclusions}
\label{sec:conclusion}

This paper generalised probabilistic modelling for comparative LLM-as-a-judge, demonstrating that existing approaches are specific instances of a broader framework. We introduced improved uncertainty estimates for individual comparisons and overall rankings, leading to more efficient iterative selection strategies. Notably, the probability of reordering proved to be a superior metric for selecting informative comparisons. We also showed the benefits of combining absolute and comparative scoring within a Product-of-Experts framework, achieving enhanced performance. While the specific expert model had limited impact on final rankings given sufficient comparisons, the choice of uncertainty estimation and the incorporation of absolute scoring significantly improved efficiency and accuracy. Our findings highlight the importance of robust uncertainty estimation in LLM-based evaluation and provide a more flexible and efficient framework for comparative assessment.

\section{Limitations}
\label{sec:limitations}

The main concern lies in the quality of the estimated uncertainties, which are crucial for the efficiency of the proposed iterative selection methods. The reliance on Laplace's approximation to derive these uncertainties introduces potential inaccuracies. This approximation assumes that the posterior distribution over model parameters is approximately Gaussian, which may not hold true in all scenarios, particularly when the true posterior is multimodal or exhibits significant skewness. Consequently, the derived uncertainty metrics, such as the variance and probability of reordering, might not perfectly reflect the true uncertainty in the model's predictions.


\newpage
\bibliography{uai2025}

\begin{thebibliography}{49}
\providecommand{\natexlab}[1]{#1}
\providecommand{\url}[1]{\texttt{#1}}
\expandafter\ifx\csname urlstyle\endcsname\relax
  \providecommand{\doi}[1]{doi: #1}\else
  \providecommand{\doi}{doi: \begingroup \urlstyle{rm}\Url}\fi

\bibitem[Achiam et~al.(2023)Achiam, Adler, Agarwal, Ahmad, Akkaya, Aleman, Almeida, Altenschmidt, Altman, Anadkat, et~al.]{achiam2023gpt}
Josh Achiam, Steven Adler, Sandhini Agarwal, Lama Ahmad, Ilge Akkaya, Florencia~Leoni Aleman, Diogo Almeida, Janko Altenschmidt, Sam Altman, Shyamal Anadkat, et~al.
\newblock Gpt-4 technical report.
\newblock \emph{arXiv preprint arXiv:2303.08774}, 2023.

\bibitem[Agresti(1990)]{agresti1990categorical}
Alan Agresti.
\newblock \emph{Categorical data analysis}.
\newblock John Wiley \& Sons, 1990.

\bibitem[Beaudoin and Swartz(2018)]{beaudoin2018computationally}
David Beaudoin and Tim Swartz.
\newblock A computationally intensive ranking system for paired comparison data.
\newblock \emph{Operations Research Perspectives}, 5:\penalty0 105--112, 2018.

\bibitem[Bradley and Terry(1952)]{bradley1952rank}
Ralph~Allan Bradley and Milton~E Terry.
\newblock Rank analysis of incomplete block designs: I. the method of paired comparisons.
\newblock \emph{Biometrika}, 39\penalty0 (3/4):\penalty0 324--345, 1952.

\bibitem[Brown et~al.(2020)Brown, Mann, Ryder, Subbiah, Kaplan, Dhariwal, Neelakantan, Shyam, Sastry, Askell, et~al.]{brown2020language}
Tom Brown, Benjamin Mann, Nick Ryder, Melanie Subbiah, Jared~D Kaplan, Prafulla Dhariwal, Arvind Neelakantan, Pranav Shyam, Girish Sastry, Amanda Askell, et~al.
\newblock Language models are few-shot learners.
\newblock \emph{Advances in neural information processing systems}, 33:\penalty0 1877--1901, 2020.

\bibitem[Bubeck et~al.(2023)Bubeck, Chandrasekaran, Eldan, Gehrke, Horvitz, Kamar, Lee, Lee, Li, Lundberg, et~al.]{bubeck2023sparks}
S{\'e}bastien Bubeck, Varun Chandrasekaran, Ronen Eldan, Johannes Gehrke, Eric Horvitz, Ece Kamar, Peter Lee, Yin~Tat Lee, Yuanzhi Li, Scott Lundberg, et~al.
\newblock Sparks of artificial general intelligence: Early experiments with gpt-4.
\newblock \emph{arXiv preprint arXiv:2303.12712}, 2023.

\bibitem[Cao et~al.(2007)Cao, Qin, Liu, Tsai, and Li]{cao2007learning}
Zhe Cao, Tao Qin, Tie-Yan Liu, Ming-Feng Tsai, and Hang Li.
\newblock Learning to rank: from pairwise approach to listwise approach.
\newblock In \emph{Proceedings of the 24th international conference on Machine learning}, pages 129--136, 2007.

\bibitem[Caron and Doucet(2012)]{caron2012efficient}
Francois Caron and Arnaud Doucet.
\newblock Efficient bayesian inference for generalized bradley--terry models.
\newblock \emph{Journal of Computational and Graphical Statistics}, 21\penalty0 (1):\penalty0 174--196, 2012.

\bibitem[Cattelan(2012)]{cattelan2012models}
Manuela Cattelan.
\newblock Models for paired comparison data: A review with emphasis on dependent data.
\newblock 2012.

\bibitem[Chen et~al.(2024{\natexlab{a}})Chen, Chen, Liu, Jiang, and Wang]{chen2024humans}
Guiming~Hardy Chen, Shunian Chen, Ziche Liu, Feng Jiang, and Benyou Wang.
\newblock Humans or llms as the judge? a study on judgement biases, 2024{\natexlab{a}}.

\bibitem[Chen et~al.(2024{\natexlab{b}})Chen, Chi, Wang, and Zhou]{chen2024premiseordermattersreasoning}
Xinyun Chen, Ryan~A. Chi, Xuezhi Wang, and Denny Zhou.
\newblock Premise order matters in reasoning with large language models, 2024{\natexlab{b}}.
\newblock URL \url{https://arxiv.org/abs/2402.08939}.

\bibitem[Chhun et~al.(2022)Chhun, Colombo, Suchanek, and Clavel]{chhun2022human}
Cyril Chhun, Pierre Colombo, Fabian Suchanek, and Chlo{\'e} Clavel.
\newblock Of human criteria and automatic metrics: A benchmark of the evaluation of story generation.
\newblock In \emph{Proceedings of the 29th International Conference on Computational Linguistics}, pages 5794--5836, 2022.

\bibitem[Chiang and Lee(2023)]{chiang2023can}
Cheng-Han Chiang and Hung-yi Lee.
\newblock Can large language models be an alternative to human evaluations?
\newblock \emph{arXiv preprint arXiv:2305.01937}, 2023.

\bibitem[Chung et~al.(2024)Chung, Hou, Longpre, Zoph, Tay, Fedus, Li, Wang, Dehghani, Brahma, et~al.]{chung2022scaling}
Hyung~Won Chung, Le~Hou, Shayne Longpre, Barret Zoph, Yi~Tay, William Fedus, Yunxuan Li, Xuezhi Wang, Mostafa Dehghani, Siddhartha Brahma, et~al.
\newblock Scaling instruction-finetuned language models.
\newblock \emph{Journal of Machine Learning Research}, 2024.

\bibitem[Csat{\'o}(2013)]{csato2013ranking}
L{\'a}szl{\'o} Csat{\'o}.
\newblock Ranking by pairwise comparisons for swiss-system tournaments.
\newblock \emph{Central European Journal of Operations Research}, 21:\penalty0 783--803, 2013.

\bibitem[David(1963)]{david1963method}
Herbert~Aron David.
\newblock \emph{The method of paired comparisons}, volume~12.
\newblock London, 1963.

\bibitem[Dubey et~al.(2024)Dubey, Jauhri, Pandey, Kadian, Al-Dahle, Letman, Mathur, Schelten, Yang, Fan, et~al.]{dubey2024llama3}
Abhimanyu Dubey, Abhinav Jauhri, Abhinav Pandey, Abhishek Kadian, Ahmad Al-Dahle, Aiesha Letman, Akhil Mathur, Alan Schelten, Amy Yang, Angela Fan, et~al.
\newblock The llama 3 herd of models.
\newblock \emph{arXiv preprint arXiv:2407.21783}, 2024.

\bibitem[Dwork et~al.(2001)Dwork, Kumar, Naor, and Sivakumar]{dwork2001rank}
Cynthia Dwork, Ravi Kumar, Moni Naor, and Dandapani Sivakumar.
\newblock Rank aggregation methods for the web.
\newblock In \emph{Proceedings of the 10th international conference on World Wide Web}, pages 613--622, 2001.

\bibitem[Fabbri et~al.(2021)Fabbri, Kry{\'s}ci{\'n}ski, McCann, Xiong, Socher, and Radev]{fabbri2021summeval}
Alexander~R Fabbri, Wojciech Kry{\'s}ci{\'n}ski, Bryan McCann, Caiming Xiong, Richard Socher, and Dragomir Radev.
\newblock Summeval: Re-evaluating summarization evaluation.
\newblock \emph{Transactions of the Association for Computational Linguistics}, 9:\penalty0 391--409, 2021.

\bibitem[Fu et~al.(2023)Fu, Ng, Jiang, and Liu]{fu2023gptscore}
Jinlan Fu, See-Kiong Ng, Zhengbao Jiang, and Pengfei Liu.
\newblock Gptscore: Evaluate as you desire.
\newblock \emph{arXiv preprint arXiv:2302.04166}, 2023.

\bibitem[Herbrich et~al.(2006)Herbrich, Minka, and Graepel]{herbrich2006trueskill}
Ralf Herbrich, Tom Minka, and Thore Graepel.
\newblock Trueskill™: a bayesian skill rating system.
\newblock \emph{Advances in neural information processing systems}, 19, 2006.

\bibitem[Hinton(1999)]{hinton1999products}
Geoffrey~E. Hinton.
\newblock Products of experts.
\newblock In \emph{Artificial Neural Networks, 1999. ICANN 99. Ninth International Conference on (Conf. Publ. No. 470)}, volume~1, pages 1--6. IET, 1999.

\bibitem[Kocmi and Federmann(2023)]{kocmi2023large}
Tom Kocmi and Christian Federmann.
\newblock Large language models are state-of-the-art evaluators of translation quality.
\newblock \emph{arXiv preprint arXiv:2302.14520}, 2023.

\bibitem[Liu et~al.(2024{\natexlab{a}})Liu, Lin, Hewitt, Paranjape, Bevilacqua, Petroni, and Liang]{liu-etal-2024-lost}
Nelson~F. Liu, Kevin Lin, John Hewitt, Ashwin Paranjape, Michele Bevilacqua, Fabio Petroni, and Percy Liang.
\newblock Lost in the middle: How language models use long contexts.
\newblock \emph{Transactions of the Association for Computational Linguistics}, 12:\penalty0 157--173, 2024{\natexlab{a}}.
\newblock \doi{10.1162/tacl_a_00638}.
\newblock URL \url{https://aclanthology.org/2024.tacl-1.9}.

\bibitem[Liu et~al.(2009)]{liu2009learning}
Tie-Yan Liu et~al.
\newblock Learning to rank for information retrieval.
\newblock \emph{Foundations and Trends{\textregistered} in Information Retrieval}, 3\penalty0 (3):\penalty0 225--331, 2009.

\bibitem[Liu et~al.(2023)Liu, Iter, Xu, Wang, Xu, and Zhu]{liu-etal-2023-g}
Yang Liu, Dan Iter, Yichong Xu, Shuohang Wang, Ruochen Xu, and Chenguang Zhu.
\newblock {G}-eval: {NLG} evaluation using gpt-4 with better human alignment.
\newblock In Houda Bouamor, Juan Pino, and Kalika Bali, editors, \emph{Proceedings of the 2023 Conference on Empirical Methods in Natural Language Processing}, pages 2511--2522, Singapore, December 2023. Association for Computational Linguistics.
\newblock \doi{10.18653/v1/2023.emnlp-main.153}.
\newblock URL \url{https://aclanthology.org/2023.emnlp-main.153}.

\bibitem[Liu et~al.(2024{\natexlab{b}})Liu, Zhou, Guo, Shareghi, Vulić, Korhonen, and Collier]{liu2024aligning}
Yinhong Liu, Han Zhou, Zhijiang Guo, Ehsan Shareghi, Ivan Vulić, Anna Korhonen, and Nigel Collier.
\newblock Aligning with human judgement: The role of pairwise preference in large language model evaluators, 2024{\natexlab{b}}.

\bibitem[Liusie et~al.(2024{\natexlab{a}})Liusie, Fathullah, and Gales]{liusie2024teacher}
Adian Liusie, Yassir Fathullah, and Mark~JF Gales.
\newblock Teacher-student training for debiasing: General permutation debiasing for large language models.
\newblock \emph{arXiv preprint arXiv:2403.13590}, 2024{\natexlab{a}}.

\bibitem[Liusie et~al.(2024{\natexlab{b}})Liusie, Manakul, and Gales]{liusie-etal-2024-llm}
Adian Liusie, Potsawee Manakul, and Mark Gales.
\newblock {LLM} comparative assessment: Zero-shot {NLG} evaluation through pairwise comparisons using large language models.
\newblock In Yvette Graham and Matthew Purver, editors, \emph{Proceedings of the 18th Conference of the European Chapter of the Association for Computational Linguistics (Volume 1: Long Papers)}, pages 139--151, St. Julian{'}s, Malta, March 2024{\natexlab{b}}. Association for Computational Linguistics.
\newblock URL \url{https://aclanthology.org/2024.eacl-long.8}.

\bibitem[Liusie et~al.(2024{\natexlab{c}})Liusie, Raina, Fathullah, and Gales]{liusie2024efficient}
Adian Liusie, Vatsal Raina, Yassir Fathullah, and Mark Gales.
\newblock Efficient llm comparative assessment: a product of experts framework for pairwise comparisons.
\newblock \emph{arXiv preprint arXiv:2405.05894}, 2024{\natexlab{c}}.

\bibitem[Louviere et~al.(2000)Louviere, Hensher, and Swait]{louviere2000stated}
Jordan~J Louviere, David~A Hensher, and Joffre~D Swait.
\newblock \emph{Stated choice methods: analysis and applications}.
\newblock Cambridge university press, 2000.

\bibitem[Manski(1977)]{manski1977structure}
Charles~F Manski.
\newblock The structure of random utility models.
\newblock \emph{Theory and decision}, 8\penalty0 (3):\penalty0 229, 1977.

\bibitem[Minka et~al.(2018)Minka, Cleven, and Zaykov]{minka2018trueskill}
Tom Minka, Ryan Cleven, and Yordan Zaykov.
\newblock Trueskill 2: An improved bayesian skill rating system.
\newblock \emph{Technical Report}, 2018.

\bibitem[Newman(2023)]{newman2023efficient}
MEJ Newman.
\newblock Efficient computation of rankings from pairwise comparisons.
\newblock \emph{Journal of Machine Learning Research}, 24\penalty0 (238):\penalty0 1--25, 2023.

\bibitem[Ouyang et~al.(2022)Ouyang, Wu, Jiang, Almeida, Wainwright, Mishkin, Zhang, Agarwal, Slama, Ray, et~al.]{ouyang2022training}
Long Ouyang, Jeffrey Wu, Xu~Jiang, Diogo Almeida, Carroll Wainwright, Pamela Mishkin, Chong Zhang, Sandhini Agarwal, Katarina Slama, Alex Ray, et~al.
\newblock Training language models to follow instructions with human feedback.
\newblock \emph{Advances in neural information processing systems}, 35:\penalty0 27730--27744, 2022.

\bibitem[Park et~al.(2024)Park, Choi, Lee, and Choo]{park2024paireval}
ChaeHun Park, Minseok Choi, Dohyun Lee, and Jaegul Choo.
\newblock Paireval: Open-domain dialogue evaluation with pairwise comparison.
\newblock \emph{arXiv preprint arXiv:2404.01015}, 2024.

\bibitem[Qin et~al.(2023)Qin, Jagerman, Hui, Zhuang, Wu, Shen, Liu, Liu, Metzler, Wang, et~al.]{qin2023large}
Zhen Qin, Rolf Jagerman, Kai Hui, Honglei Zhuang, Junru Wu, Jiaming Shen, Tianqi Liu, Jialu Liu, Donald Metzler, Xuanhui Wang, et~al.
\newblock Large language models are effective text rankers with pairwise ranking prompting.
\newblock \emph{arXiv preprint arXiv:2306.17563}, 2023.

\bibitem[{Qwen Team}(2024)]{qwen2.5}
{Qwen Team}.
\newblock Qwen2.5: A party of foundation models, September 2024.
\newblock URL \url{https://qwenlm.github.io/blog/qwen2.5/}.

\bibitem[Raina et~al.(2024)Raina, Liusie, and Gales]{raina2024finetuningllmscomparativeassessment}
Vatsal Raina, Adian Liusie, and Mark Gales.
\newblock Finetuning llms for comparative assessment tasks, 2024.
\newblock URL \url{https://arxiv.org/abs/2409.15979}.

\bibitem[Taori et~al.(2023)Taori, Gulrajani, Zhang, Dubois, Li, Guestrin, Liang, and Hashimoto]{taori2023stanford}
Rohan Taori, Ishaan Gulrajani, Tianyi Zhang, Yann Dubois, Xuechen Li, Carlos Guestrin, Percy Liang, and Tatsunori~B Hashimoto.
\newblock Stanford alpaca: An instruction-following llama model, 2023.

\bibitem[Wang et~al.(2023{\natexlab{a}})Wang, Liang, Meng, Shi, Li, Xu, Qu, and Zhou]{wang2023chatgpt}
Jiaan Wang, Yunlong Liang, Fandong Meng, Haoxiang Shi, Zhixu Li, Jinan Xu, Jianfeng Qu, and Jie Zhou.
\newblock Is chatgpt a good nlg evaluator? a preliminary study.
\newblock \emph{arXiv preprint arXiv:2303.04048}, 2023{\natexlab{a}}.

\bibitem[Wang et~al.(2023{\natexlab{b}})Wang, Li, Chen, Cai, Zhu, Lin, Cao, Liu, Liu, and Sui]{wang2023large}
Peiyi Wang, Lei Li, Liang Chen, Zefan Cai, Dawei Zhu, Binghuai Lin, Yunbo Cao, Qi~Liu, Tianyu Liu, and Zhifang Sui.
\newblock Large language models are not fair evaluators, 2023{\natexlab{b}}.

\bibitem[Wang et~al.(2023{\natexlab{c}})Wang, Cai, Chen, Liang, and Hooi]{wang-etal-2023-primacy}
Yiwei Wang, Yujun Cai, Muhao Chen, Yuxuan Liang, and Bryan Hooi.
\newblock Primacy effect of {C}hat{GPT}.
\newblock In Houda Bouamor, Juan Pino, and Kalika Bali, editors, \emph{Proceedings of the 2023 Conference on Empirical Methods in Natural Language Processing}, pages 108--115, Singapore, December 2023{\natexlab{c}}. Association for Computational Linguistics.
\newblock \doi{10.18653/v1/2023.emnlp-main.8}.
\newblock URL \url{https://aclanthology.org/2023.emnlp-main.8}.

\bibitem[Wang et~al.(2022)Wang, Kordi, Mishra, Liu, Smith, Khashabi, and Hajishirzi]{wang2022self}
Yizhong Wang, Yeganeh Kordi, Swaroop Mishra, Alisa Liu, Noah~A Smith, Daniel Khashabi, and Hannaneh Hajishirzi.
\newblock Self-instruct: Aligning language models with self-generated instructions.
\newblock \emph{arXiv preprint arXiv:2212.10560}, 2022.

\bibitem[Welling(2007)]{Welling:2007}
M.~Welling.
\newblock {P}roduct of experts.
\newblock \emph{Scholarpedia}, 2\penalty0 (10):\penalty0 3879, 2007.
\newblock \doi{10.4249/scholarpedia.3879}.
\newblock revision \#137078.

\bibitem[Zermelo(1929)]{zermelo1929berechnung}
Ernst Zermelo.
\newblock Die berechnung der turnier-ergebnisse als ein maximumproblem der wahrscheinlichkeitsrechnung.
\newblock \emph{Mathematische Zeitschrift}, 29\penalty0 (1):\penalty0 436--460, 1929.

\bibitem[Zheng et~al.(2023)Zheng, Chiang, Sheng, Zhuang, Wu, Zhuang, Lin, Li, Li, Xing, et~al.]{zheng2023judging}
Lianmin Zheng, Wei-Lin Chiang, Ying Sheng, Siyuan Zhuang, Zhanghao Wu, Yonghao Zhuang, Zi~Lin, Zhuohan Li, Dacheng Li, Eric Xing, et~al.
\newblock Judging llm-as-a-judge with mt-bench and chatbot arena.
\newblock \emph{arXiv preprint arXiv:2306.05685}, 2023.

\bibitem[Zhou et~al.(2023)Zhou, Liu, Xu, Iyer, Sun, Mao, Ma, Efrat, Yu, YU, Zhang, Ghosh, Lewis, Zettlemoyer, and Levy]{lima2023zhou}
Chunting Zhou, Pengfei Liu, Puxin Xu, Srinivasan Iyer, Jiao Sun, Yuning Mao, Xuezhe Ma, Avia Efrat, Ping Yu, LILI YU, Susan Zhang, Gargi Ghosh, Mike Lewis, Luke Zettlemoyer, and Omer Levy.
\newblock Lima: Less is more for alignment.
\newblock In A.~Oh, T.~Naumann, A.~Globerson, K.~Saenko, M.~Hardt, and S.~Levine, editors, \emph{Advances in Neural Information Processing Systems}, volume~36, pages 55006--55021. Curran Associates, Inc., 2023.
\newblock URL \url{https://proceedings.neurips.cc/paper_files/paper/2023/file/ac662d74829e4407ce1d126477f4a03a-Paper-Conference.pdf}.

\bibitem[Zhu et~al.(2023)Zhu, Wang, and Wang]{zhu2023judgelm}
Lianghui Zhu, Xinggang Wang, and Xinlong Wang.
\newblock Judgelm: Fine-tuned large language models are scalable judges.
\newblock 2023.

\end{thebibliography}

\newpage
\onecolumn
\title{Generalised Probabilistic Modelling and Improved Uncertainty Estimation in Comparative LLM-as-a-judge\\(Supplementary Material)}
\maketitle

\appendix
\section{Prompting}
\label{app:prompting}

When prompting an LLM for the score of a candidate, or which of two candidates is better, the important information lies in the logits of the tokens we are interested in. In this section we detail the design the of our prompts for both Flan-T5 and Qwen2.5. Since the former is an encoder-decoder foundation model and the latter is a decoder-only foundation model the prompts need to be designed slightly differently.

\textbf{Absolute prompting}: For the Flan-T5 system, we give the encoder the following prompt:
\begin{align*}
    & \text{\tt Article: <context>\n\n Summary: <A>}\\
    & \text{\tt \n\n Score the response between 1 }\\
    & \text{\tt and 10 based on how coherent the} \\
    & \text{\tt summary is.}
\end{align*}
where we are scoring the coherency of a summary. The {\tt <context>} and {\tt <A>} are replaced by the article and summary. Following this we extract the logits corresponding to 1 to 10 from the decoder. The probability of each class is then:
\begin{align*}
    p_c = \frac{\exp(z_c)}{\sum_{i = 1}^{10} \exp(z_i)} \hspace{2mm} c = 1, \dots, 10
\end{align*}
The choice of 1-10 is arbitrary and any other range could have been chosen.

\textbf{Comparative prompting}: For the Flan-T5 system, we give the encoder the following prompt:
\begin{align*}
    & \text{\tt Article: <context>\n\n Summary A: <A>}\\
    & \text{\tt \n\n Summary B: <B>\n\n Which Summary}\\
    & \text{\tt is more coherent, Summary A or}\\
    & \text{\tt Summary B?}
\end{align*}
The {\tt <context>}, {\tt <A>} and {\tt <B>} are replaced by the article and two different summaries. Following this we give the following prefix to the decoder:
\begin{align*}
    & \text{\tt Summary }
\end{align*}
and extract the logits corresponding to A and B from the decoder. The prefix ensures that the probability mass of the next token is concentrated into the options "A" and "B". From these logits we extract the probability that A will win:
\begin{align*}
    p = \frac{\exp(z_A)}{\exp(z_A) + \exp(z_B)}
\end{align*}

Similarly, we prompt the Qwen2.5 system in the following matter when we want to rank various stories from HANNA:
\begin{align*}
    & \text{\tt \{"role": "system", "content": "You} \\
    & \text{\tt are an expert story assessor."\},} \\\\
    & \text{\tt \{"role": "user", "content": "Story A:} \\
    & \text{\tt <A>\n\n Story B: <B>\n\n Which story} \\
    & \text{\tt is better overall, Story A or B?} \\
    & \text{\tt Answer only with Story A or Story B."\}} \\\\
    & \text{\tt \{"role": "assistant", "content": "Story "\}}
\end{align*}
These are then prepared by the Qwen2.5 tokenizer in the instruction following format and fed into the model. Following on, the logits corresponding to $A$ and $B$ are then extracted for the next token and converted into a probability.

\section{Calibration}
\label{app:calibration}

This section reports the calibration error and reliability diagrams for the different metrics under a biased and debiased setup. The main point is to address the overconfidence issue related to our results in Section \ref{ssec:iterative} and why temperature annealing is not enough to solve the problem.

\begin{figure*}[h!]
    \centering
    \begin{subfigure}{0.495\textwidth}
        \includegraphics[width=\textwidth]{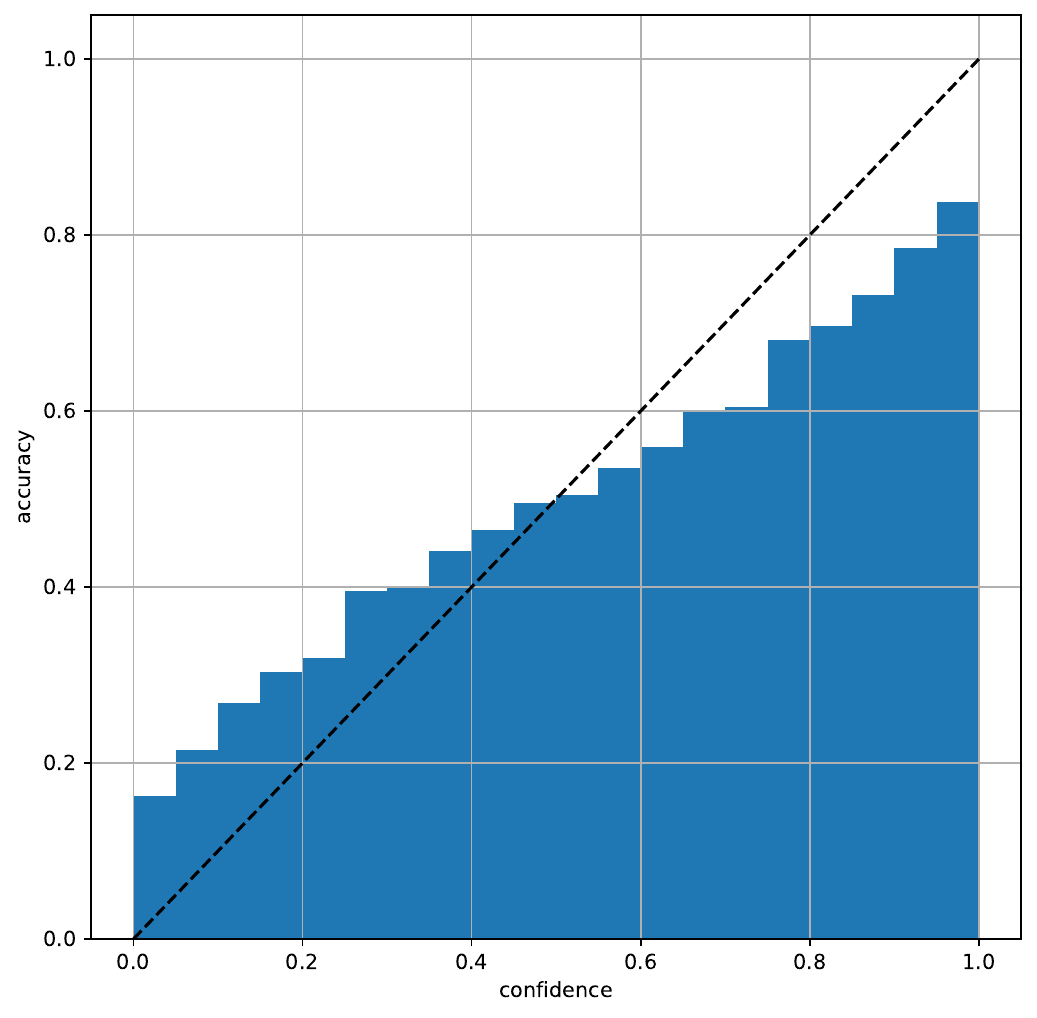}
        \caption{Biased}
        \label{fig:cal-biased}
    \end{subfigure}
    \hfill
    \begin{subfigure}{0.495\textwidth}
        \includegraphics[width=\textwidth]{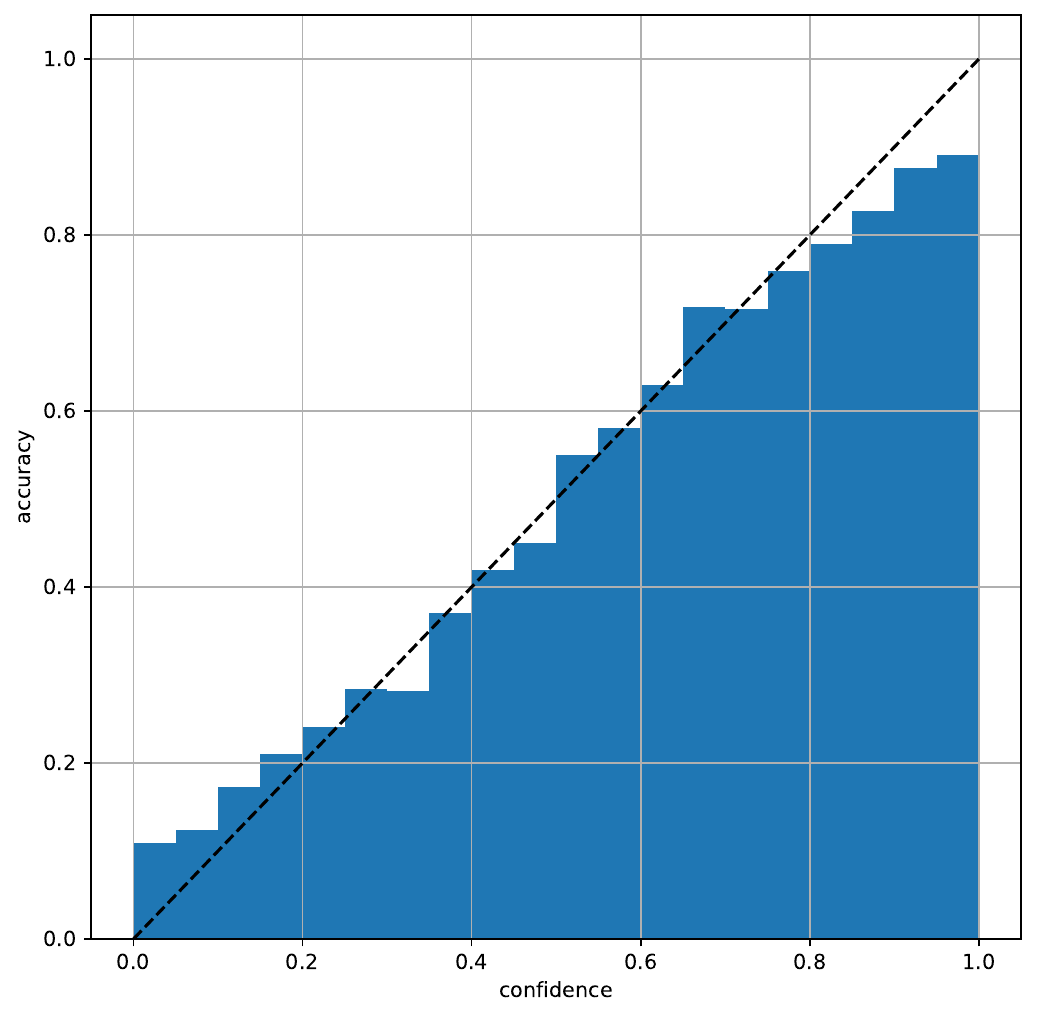}
        \caption{Debiased}
        \label{fig:cal-debiased}
    \end{subfigure}
    \hfill
    \begin{subfigure}{0.495\textwidth}
        \includegraphics[width=\textwidth]{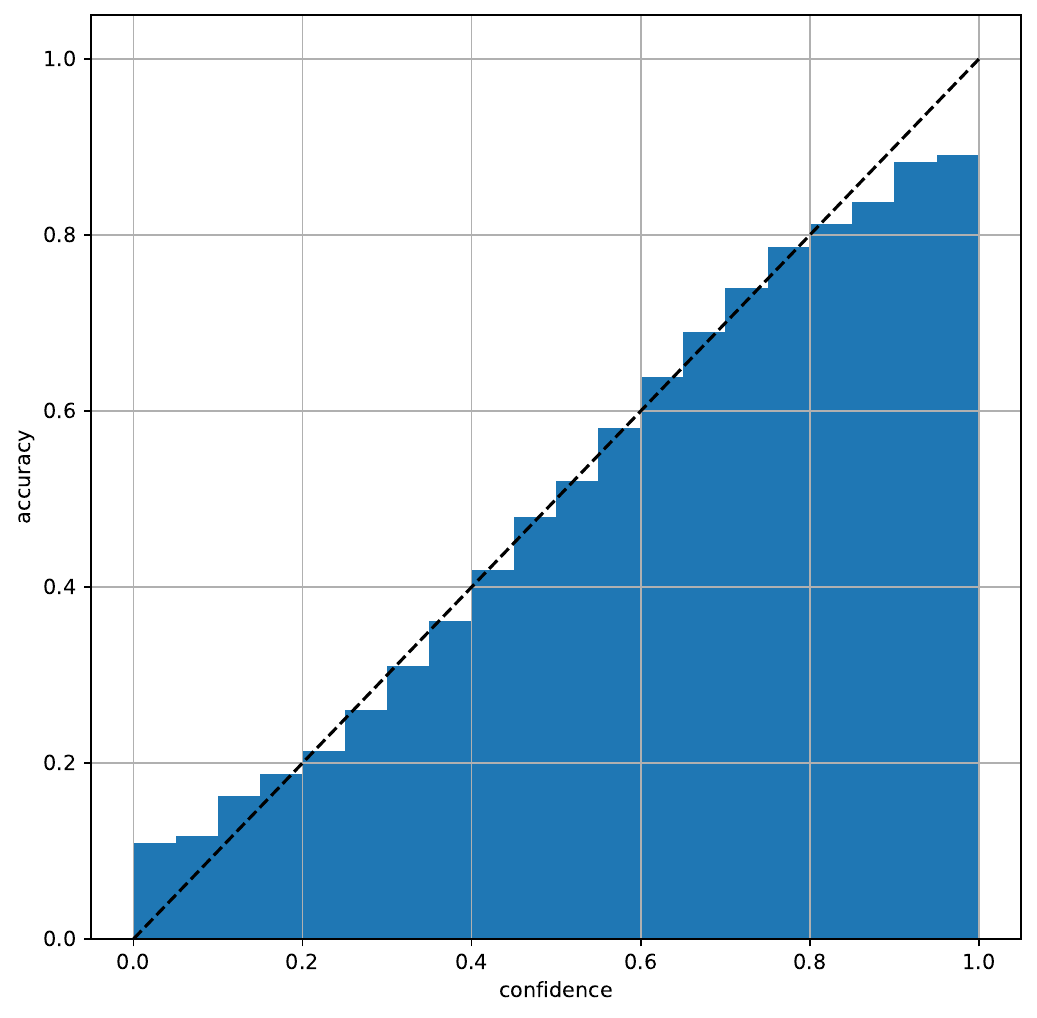}
        \caption{\tt Biased - calibrated}
        \label{fig:cal-biased-calibrated}
    \end{subfigure}
    \hfill
    \begin{subfigure}{0.495\textwidth}
        \includegraphics[width=\textwidth]{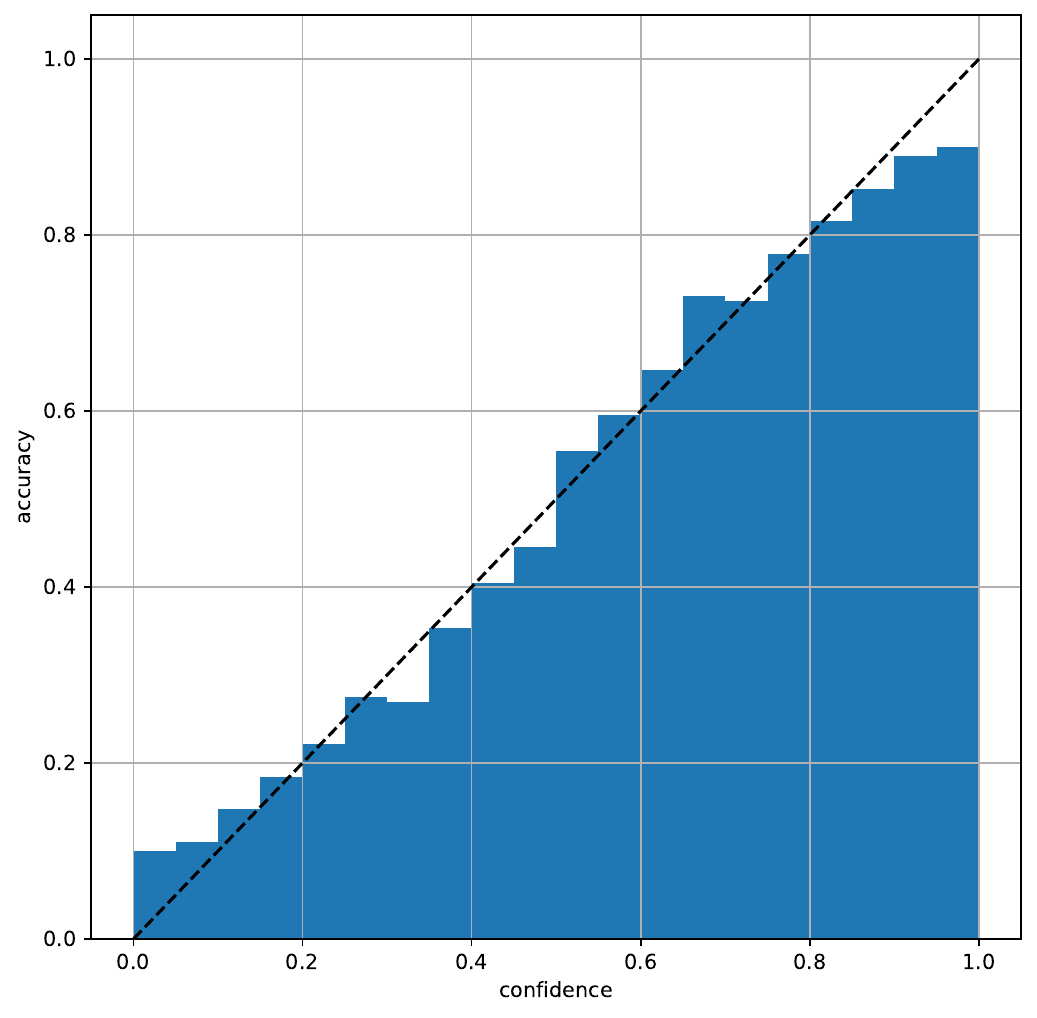}
        \caption{\tt Deiased - calibrated}
        \label{fig:cal-debiased-calibrated}
    \end{subfigure}
    \caption{The reliability diagram of biased and debiased, standard and calibrated systems on the \texttt{coherency} metric.}
    \label{fig:relib}
\end{figure*}

The calibration is based on the confidence scores of individual comparisons $\max(p, 1 - p)$. Therefore, when calibrating using temperature annealing, the resulting (binary) predictions remain the same:
\begin{align*}
    \tilde{p} = \frac{p^{1/T}}{p^{1/T} + (1-p)^{1/T}}
\end{align*}
To understand the impact of calibration, we find the optimal temperature on the SummEval dataset by minimising the expected calibration error, see Table \ref{tab:calib}. Each attribute has its optimal temperature.
\begin{table}[h!]
	\centering{}
            \small
		\begin{center}
                \caption{Flan-T5 (3B): Expected calibration error (\%).}
			\vspace{-2mm}
			\def\arraystretch{1.0}
			\begin{adjustbox}{center}
    			\begin{tabular}{c|c|cccc}
    				\toprule
                        Method & Debiased & {\tt COH} & {\tt CON} & {\tt FLU} & {\tt REL} \\
                        \midrule
                        \multirow{2}{*}{-} 
                        & \xmark & 9.80 & 3.77 & 9.87 & 11.86 \\
                        & \cmark & 2.83 & 1.82 & 4.84 & 6.20 \\
                        \midrule
                        \multirow{2}{*}{Calibrated} 
                        & \xmark & 1.02 & 0.68 & 1.28 & 0.98 \\
                        & \cmark & 2.58 & 1.72 & 1.08 & 1.07 \\
    				\bottomrule 
    			\end{tabular}
    		\end{adjustbox}
			\label{tab:calib}
		\end{center}
\end{table}
We also report the corresponding reliability diagrams in Figure \ref{fig:relib}. From these results, it is evident that simple temperature annealing can almost entirely resolve the miscalibration in the systems.

This next part will check how temperature annealing affects the solution of a soft Bradley-Terry model. Starting from the gradient of the log-likelihood:
\begin{align*}
    \nabla \ln\tp(\bm s \vert \mathcal{C}_{1:K}) =\hspace{-2mm}\sum_{i, j \in \mathcal{C}_{1:K}} p_{ij} - \sigma(s_i - s_j) = 0
\end{align*}
Looking at a single element of the sum, and under calibrated probabilities the new solution becomes:
\begin{align*}
    \tilde{p}_{ij} = \sigma(\tilde{s}_i - \tilde{s}_j) &\iff \\
    \frac{p^{1/T}}{p^{1/T} + (1-p)^{1/T}} = \sigma(\tilde{s}_i - \tilde{s}_j) &\iff \\
    \frac{1}{1 + \left(\frac{1-p}{p}\right)^{1/T}} = \sigma(\tilde{s}_i - \tilde{s}_j) &\iff \\    
    \frac{1}{1 + \exp\left(\frac{1}{T}\ln\left(\frac{1-p}{p}\right)\right)} = \sigma(\tilde{s}_i - \tilde{s}_j) &\iff
\end{align*}
\begin{align*}
    \sigma\left(\frac{1}{T}\ln\left(\frac{p}{1-p}\right)\right) = \sigma(\tilde{s}_i - \tilde{s}_j) &\iff \\    
    \frac{1}{T}\ln\left(\frac{p}{1-p}\right) = \tilde{s}_i - \tilde{s}_j &\iff \\    
    \frac{p}{1-p} = \exp(T(\tilde{s}_i - \tilde{s}_j)) &\iff \\    
    p = \frac{\exp(T(\tilde{s}_i - \tilde{s}_j))}{1 + \exp(T(\tilde{s}_i - \tilde{s}_j))} &\iff \\    
    p = \frac{1}{1 + \exp(-T(\tilde{s}_i - \tilde{s}_j))} &\iff \\    
    p = \sigma(T(\tilde{s}_i - \tilde{s}_j)) & 
\end{align*}
This shows that temperature annealing leads to a new solution of scores that are linearly scaled by the temperature $T$. Therefore, even if temperature annealing is enough to calibrate a system, it has no impact at all on the predicted rankings. 

\begin{figure*}[h!]
    \centering
    \begin{subfigure}{0.495\textwidth}
        \includegraphics[width=\textwidth]{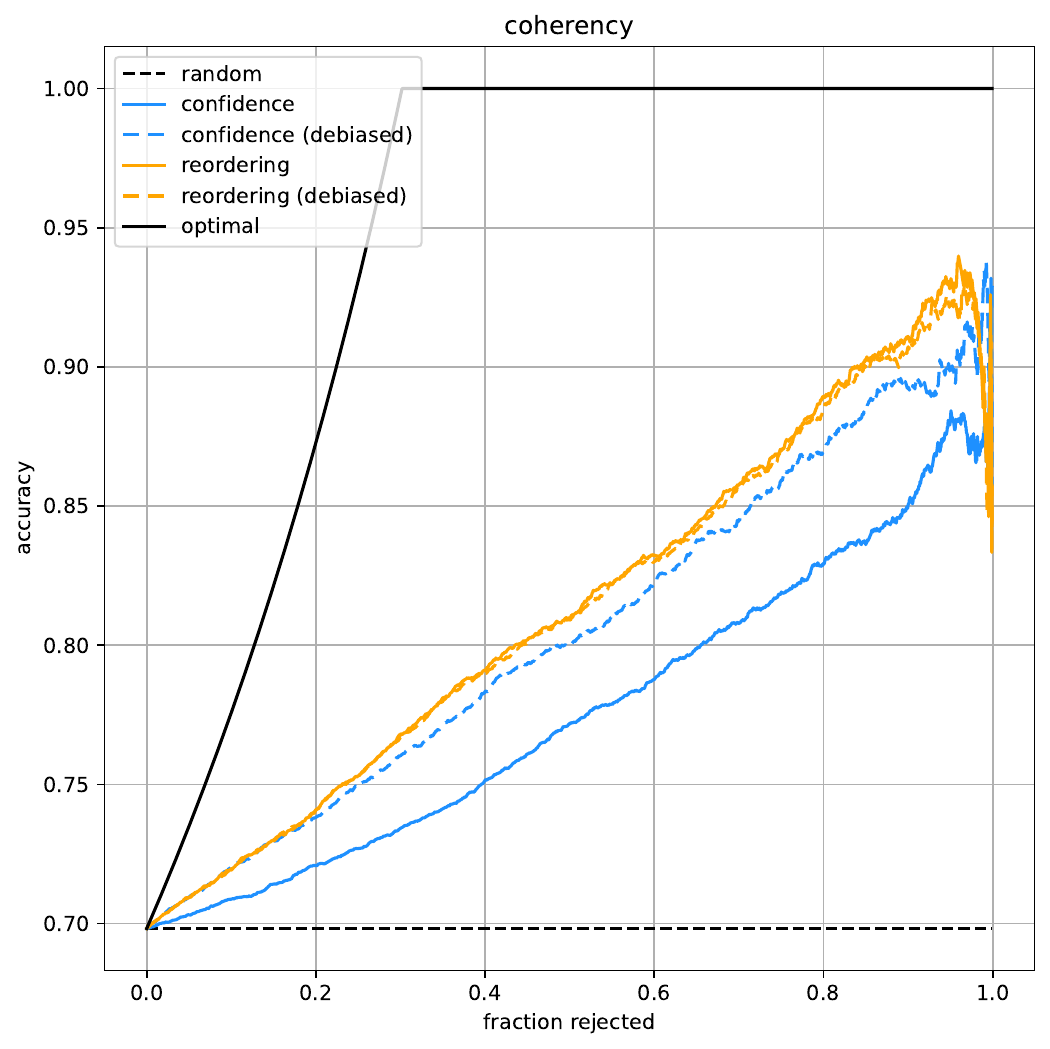}
        \caption{\tt COH}
        \label{fig:acc-coh}
    \end{subfigure}
    \hfill
    \begin{subfigure}{0.495\textwidth}
        \includegraphics[width=\textwidth]{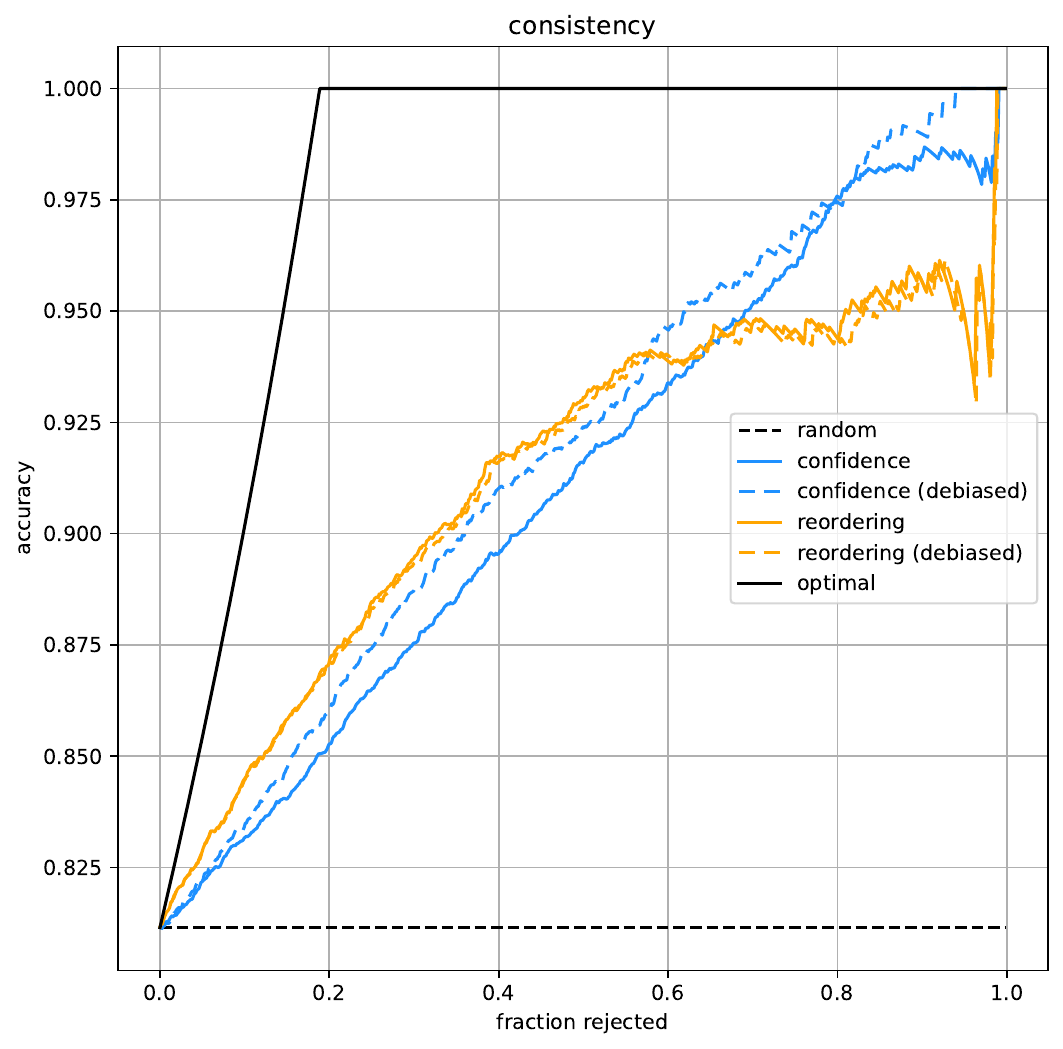}
        \caption{\tt CON}
        \label{fig:acc-con}
    \end{subfigure}
    \hfill
    \begin{subfigure}{0.495\textwidth}
        \includegraphics[width=\textwidth]{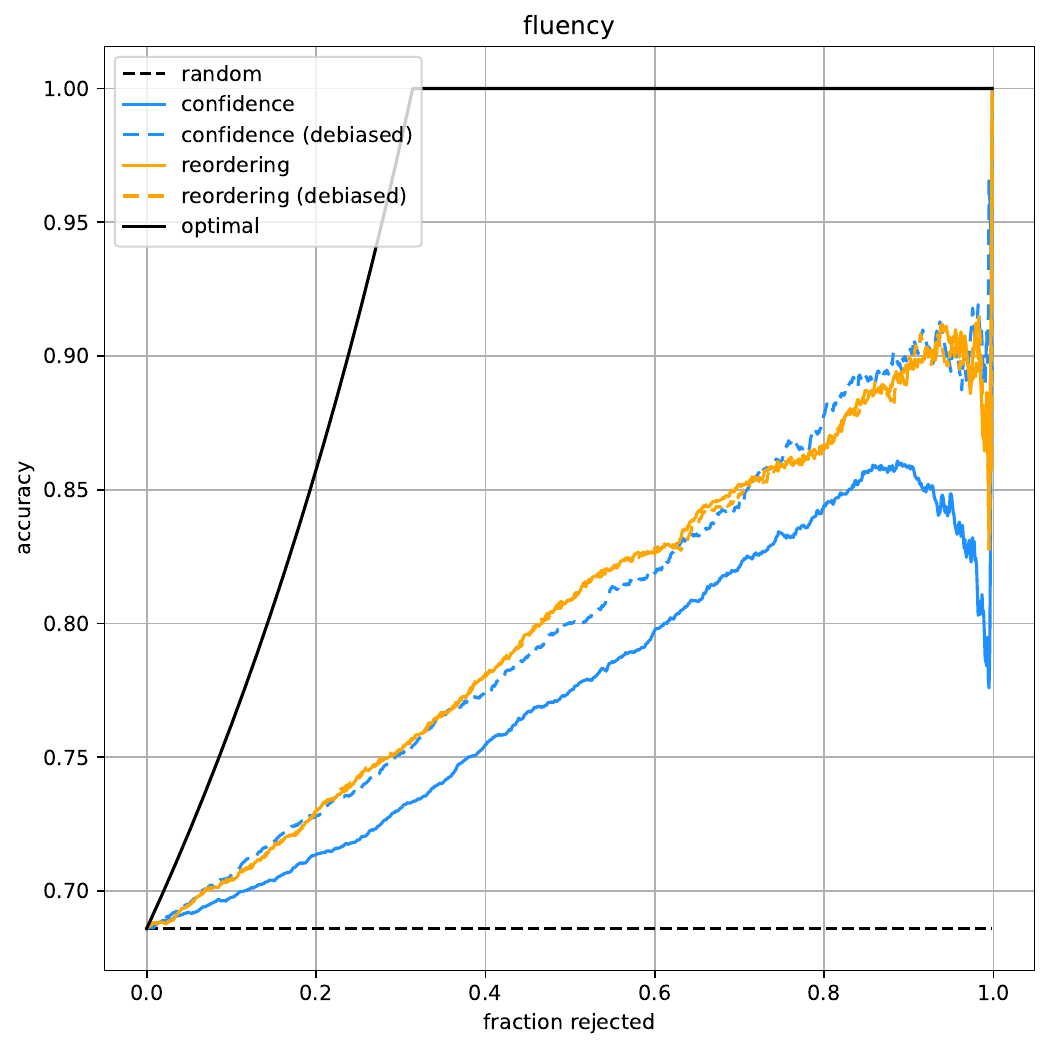}
        \caption{\tt FLU}
        \label{fig:acc-flu}
    \end{subfigure}
    \hfill
    \begin{subfigure}{0.495\textwidth}
        \includegraphics[width=\textwidth]{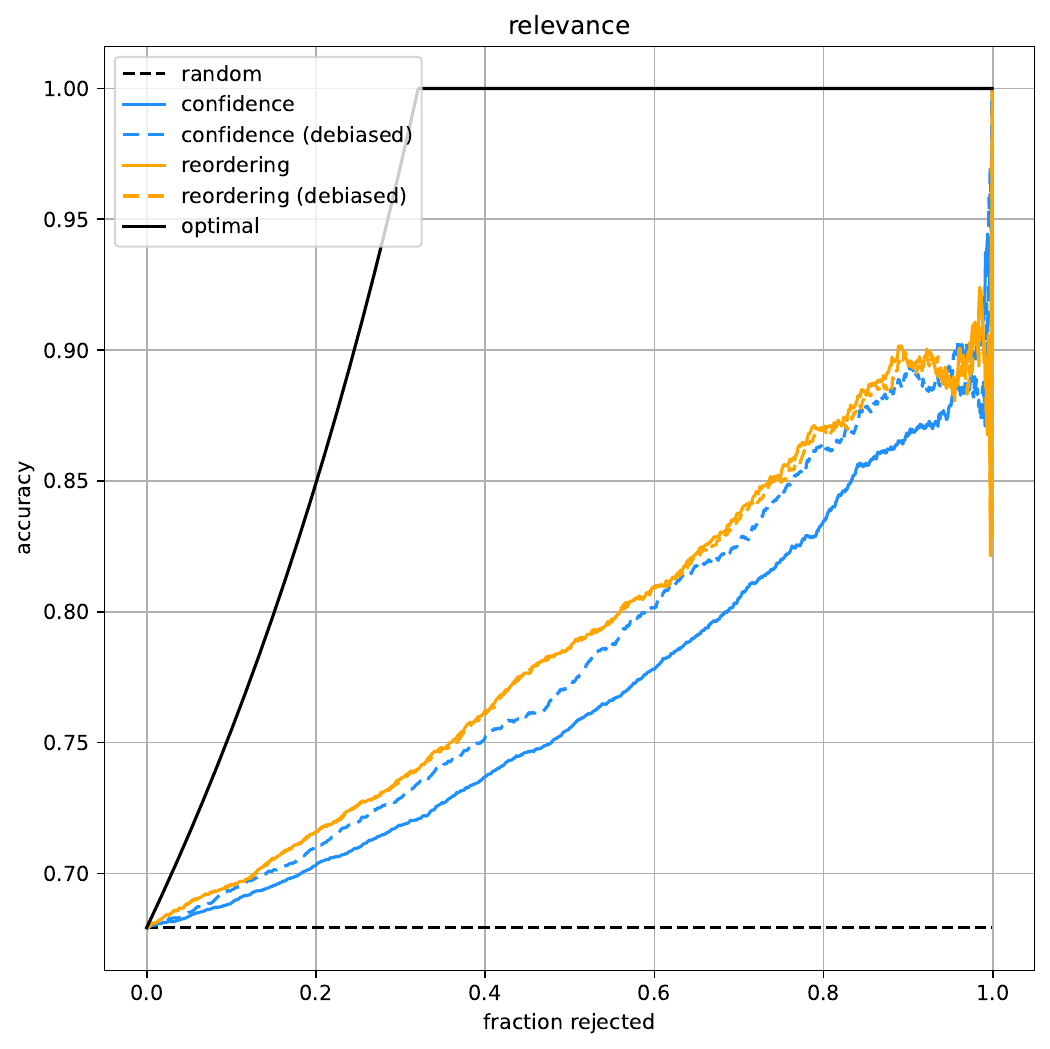}
        \caption{\tt REL}
        \label{fig:acc-rl}
    \end{subfigure}
    \caption{The accuracy at a comparison-level when the examples of lowest confidence/highest uncertainty are rejected.}
    \label{fig:rejection}
\end{figure*}

Instead, we report a different result, the diagrams in Figure \ref{fig:rejection}. We obtain either the confidence of each comparison or the probability of reordering. Then the aim is to compute the accuracy of comparisons on a filtered dataset when removing the examples of lowest confidence/highest uncertainty. 
What one expects from high quality uncertainties is for the accuracy of the filtered dataset to improve as much as possible. While we observe that the accuracy improves as we reject samples, both metrics display a significant overconfidence issue; accuracy reduces when rejecting samples with the highest confidence and lowest uncertainty. 
This could partially explain why our results in Section \ref{ssec:iterative} showcase a 'bump', where adding more comparisons decreases the system's performance. This also justifies using more advanced methods for calibrating the outputs of LLM-judges when using them to rank candidates.

\begin{figure*}[h!]
    \centering
    \begin{subfigure}{0.495\textwidth}
        \includegraphics[width=\textwidth]{Figures/calibration/calibration-coherency-original-symmetric-False.pdf}
        \vspace{-6mm}
        \caption{Flan-T5 (3B): ECE = 9.80\%}
        \vspace{2mm}
        \label{fig:cal-biased-flan}
    \end{subfigure}
    \hfill
    \begin{subfigure}{0.495\textwidth}
        \includegraphics[width=\textwidth]{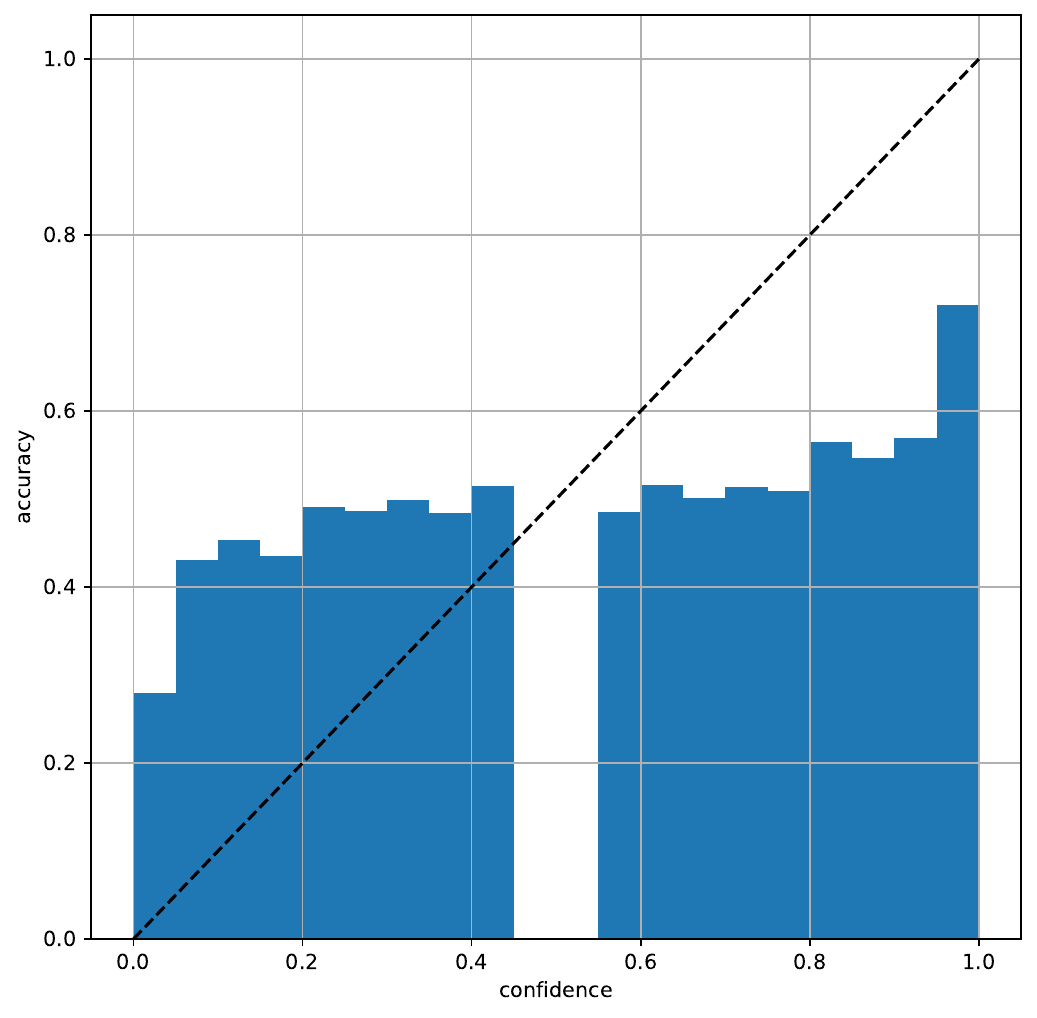}
        \vspace{-6mm}
        \caption{Qwen2.5-Instruct (3B): ECE = 25.73\%}
        \vspace{2mm}
        \label{fig:cal-biased-qwen-3b}
    \end{subfigure}
    \hfill
    \begin{subfigure}{0.495\textwidth}
        \includegraphics[width=\textwidth]{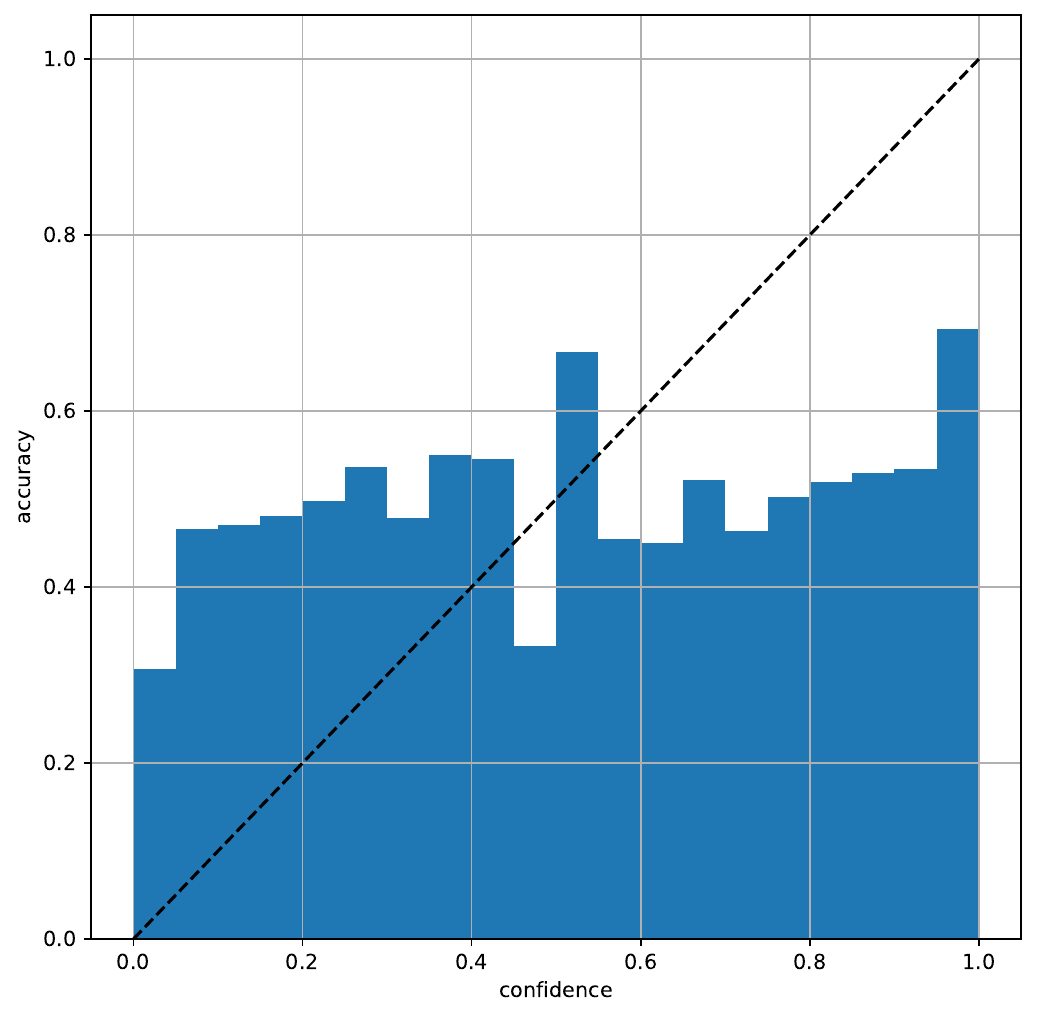}
        \vspace{-6mm}
        \caption{Qwen2.5-Instruct (7B): ECE = 28.17\%}
        \label{fig:cal-biased-qwen-7b}
    \end{subfigure}
    \hfill
    \begin{subfigure}{0.495\textwidth}
        \includegraphics[width=\textwidth]{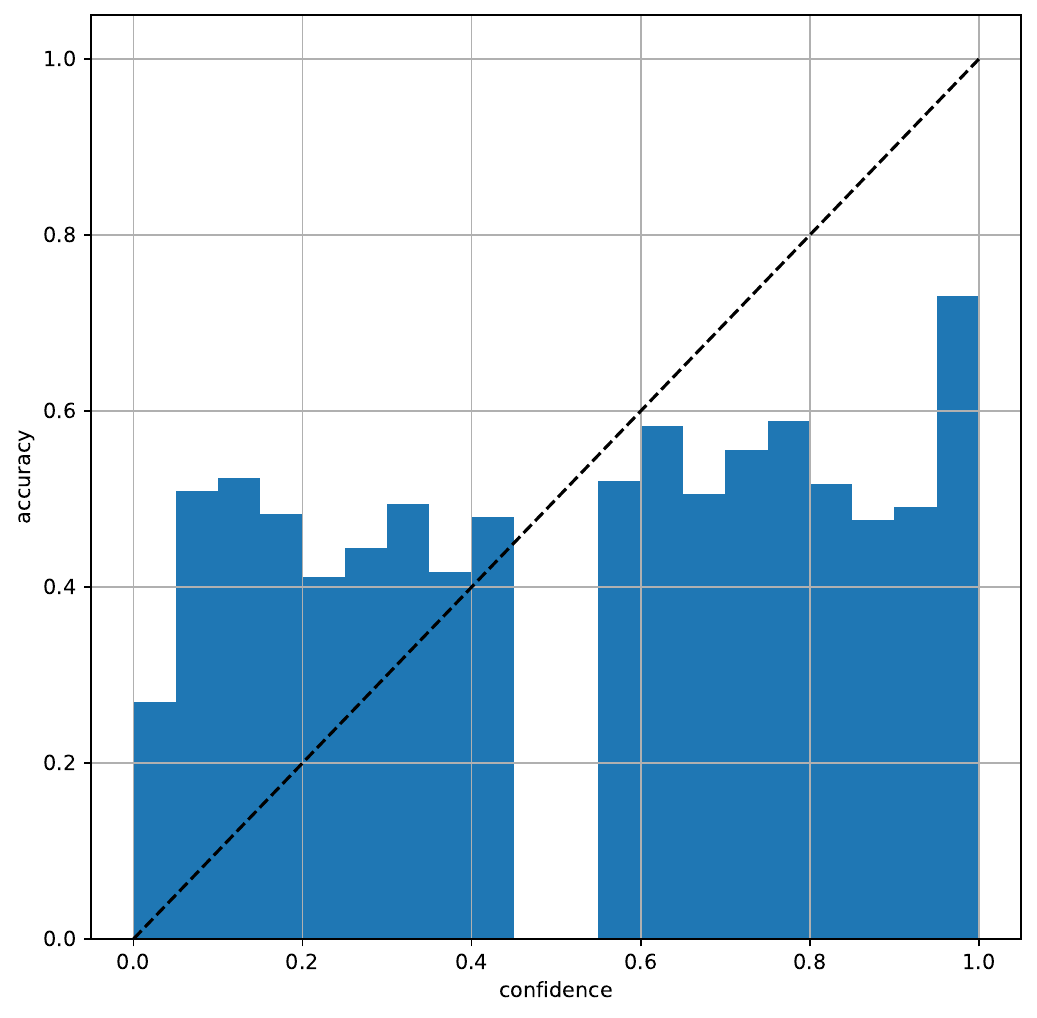}
        \vspace{-6mm}
        \caption{Qwen2.5-Instruct (14B): ECE = 24.64\%}
        \label{fig:cal-biased-qwen-14b}
    \end{subfigure}
    \caption{The reliability diagram of biased systems on the \texttt{coherency} metric. The expected calibration error (ECE) is reported for each system.}
    \label{fig:relib-all}
\end{figure*}

Finally, we report the reliability diagram of all language models on SummEval ({\tt COH}), see Figure \ref{fig:relib-all}. While the Qwen2.5 family performs better in predicting the correct ranking, these systems suffer more from overconfidence as evidenced by the much higher calibration error. As discussed, temperature annealing can resolve the calibration issue without having any impact on performance. Instead, viable approaches that could resolve this include:
\begin{enumerate}
    \item Ensembling multiple comparative LLM-as-a-judge backbones. Averaging outputs is known to reduce calibration error and the overconfidence issue.
    \item Ensembling multiple prompts using the same LLM backbone. Alternatively to above, one can design multiple prompts which achieve the same task, and average the output from each one. 
\end{enumerate}
In addition to the above, one can aim to design improved prompts which achieve the task and minimise biases in the system. We leave these approaches to future work.

\clearpage

\clearpage

\clearpage
\section{Probability of Reordering}
\label{app:reordering}
In this section we showcase how the probability of reordering can be rephrased to a familiar form. Starting from the general Laplace's approximation:
\begin{align*}
    \tp(\bm s \vert\mathcal{C}_{1:K}) = \mathcal{N}(\bm s ; \bm\mu, \bm\Sigma)
\end{align*}
the distribution of a pair of candidates becomes:
\begin{align*}
    \tp\!\left(s_i, s_j \vert\mathcal{C}_{1:K}\right) 
    = \mathcal{N}\left(
    \begin{bmatrix}
           s_{i} \\
           s_{j}
    \end{bmatrix} ;
    \begin{bmatrix}
           \mu_{i} \\
           \mu_{j}
    \end{bmatrix}, 
    \begin{bmatrix}
           \Sigma_{ii} & \Sigma_{ij} \\
           \Sigma_{ji} & \Sigma_{jj}
    \end{bmatrix}
    \right)
\end{align*}
We are only interested in the distribution of the difference $s_i - s_j$:
\begin{align*}
    \tp\!\left(s_i - s_j \vert\mathcal{C}_{1:K}\right) 
    = \mathcal{N}\left(
    s_i - s_j ;
    \mu_i - \mu_j, 
    \Sigma_{ii} - 2\Sigma_{ij} + \Sigma_{jj}
    \right)
\end{align*}
Assuming that $s_i > s_j$ ($\mu_i > \mu_j$) the probability of reordering becomes:
\begin{align*}
    \tP(s_i < s_j\vert\mathcal{C}_{1:K}) 
    = \Phi\left( \frac{\mu_j - \mu_i}{\sqrt{\Sigma_{ii} - 2\Sigma_{ij} + \Sigma_{jj}}}  \right)
\end{align*}
The selection is based on picking the examples with highest probability of reordering:
\begin{align*}
    \hat{i}, \hat{j} 
    & = \argmax_{i, j} 
    \Phi\left( \frac{\mu_j - \mu_i}{\sqrt{\Sigma_{ii} - 2\Sigma_{ij} + \Sigma_{jj}}}  \right) \\
    & = \argmax_{i, j} \frac{\mu_j - \mu_i}{\sqrt{\Sigma_{ii} - 2\Sigma_{ij} + \Sigma_{jj}}} \\
    & = \argmin_{i, j} -\frac{\sqrt{\Sigma_{ii} - 2\Sigma_{ij} + \Sigma_{jj}}}{\mu_i - \mu_j} \\
    & = \argmax_{i, j} \frac{\Sigma_{ii} - 2\Sigma_{ij} + \Sigma_{jj}}{(\mu_i - \mu_j)^2}
\end{align*}
Similarly, assuming $s_j > s_i$ returns the exact same expression.

\clearpage
\section{Extended Results for Qwen Systems}
\label{app:results}

\begin{figure*}[h!]
    \begin{subfigure}{0.49\textwidth}
        \includegraphics[width=\textwidth]{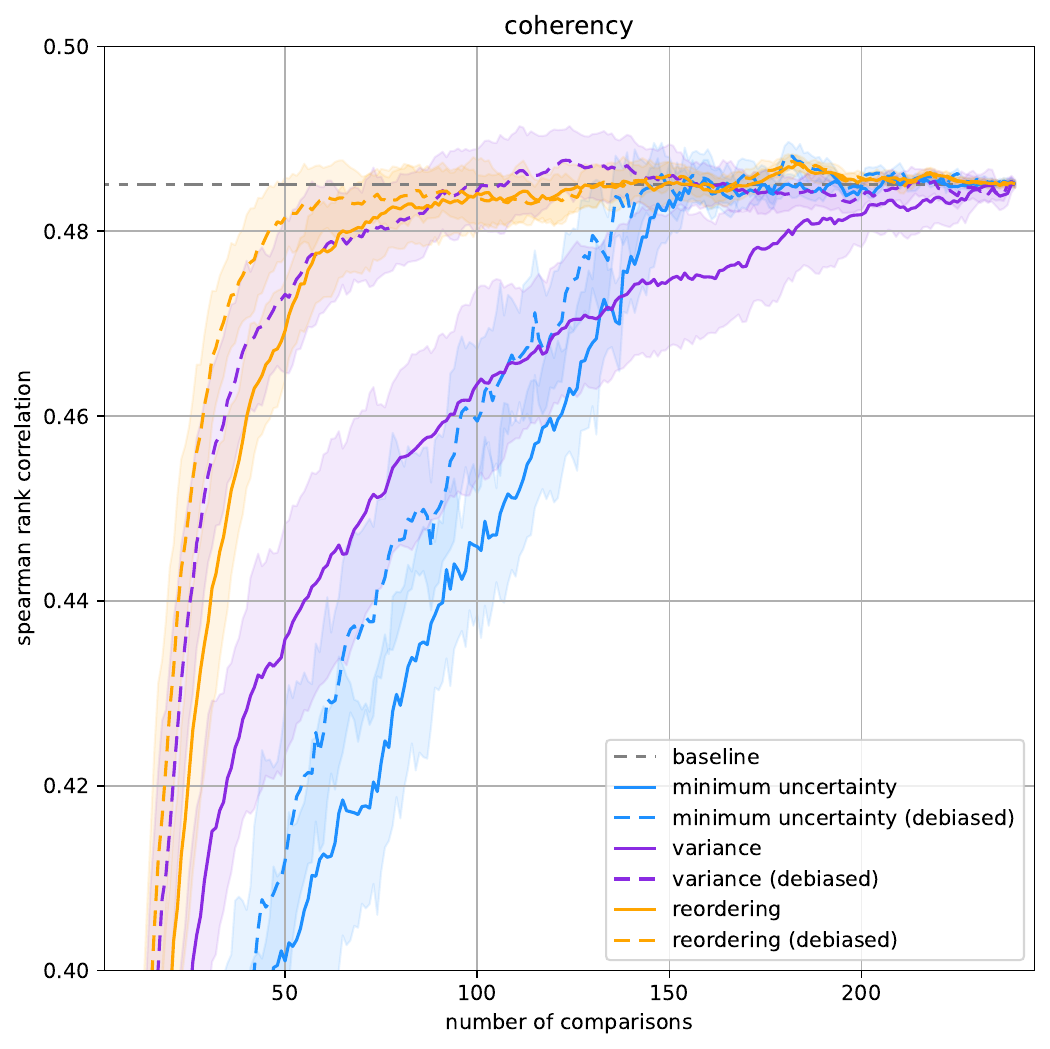}
        \label{fig:sel-coh}
    \end{subfigure}
    \hfill
    \begin{subfigure}{0.49\textwidth}
        \includegraphics[width=\textwidth]{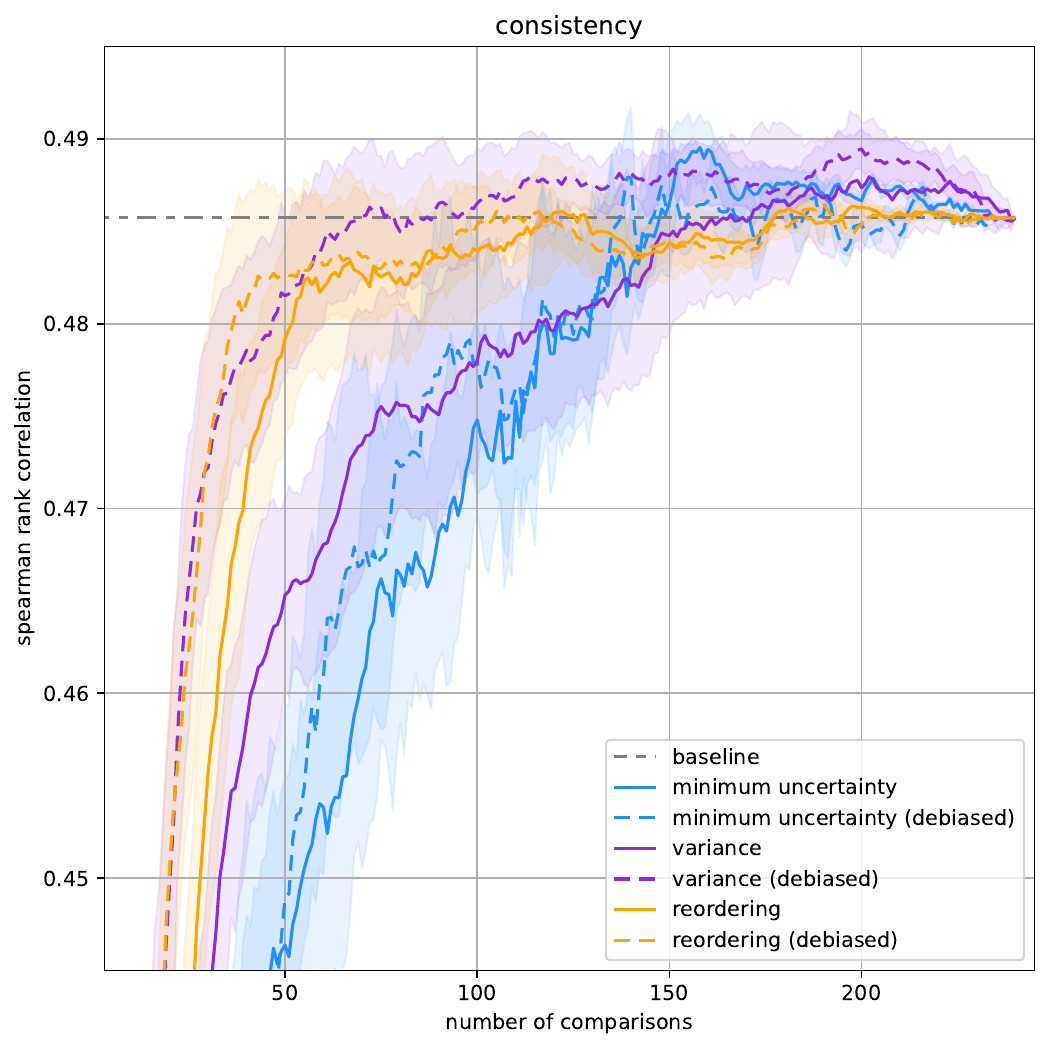}
        \label{fig:sel-con}
    \end{subfigure}
    \hfill
    \begin{subfigure}{0.49\textwidth}
        \includegraphics[width=\textwidth]{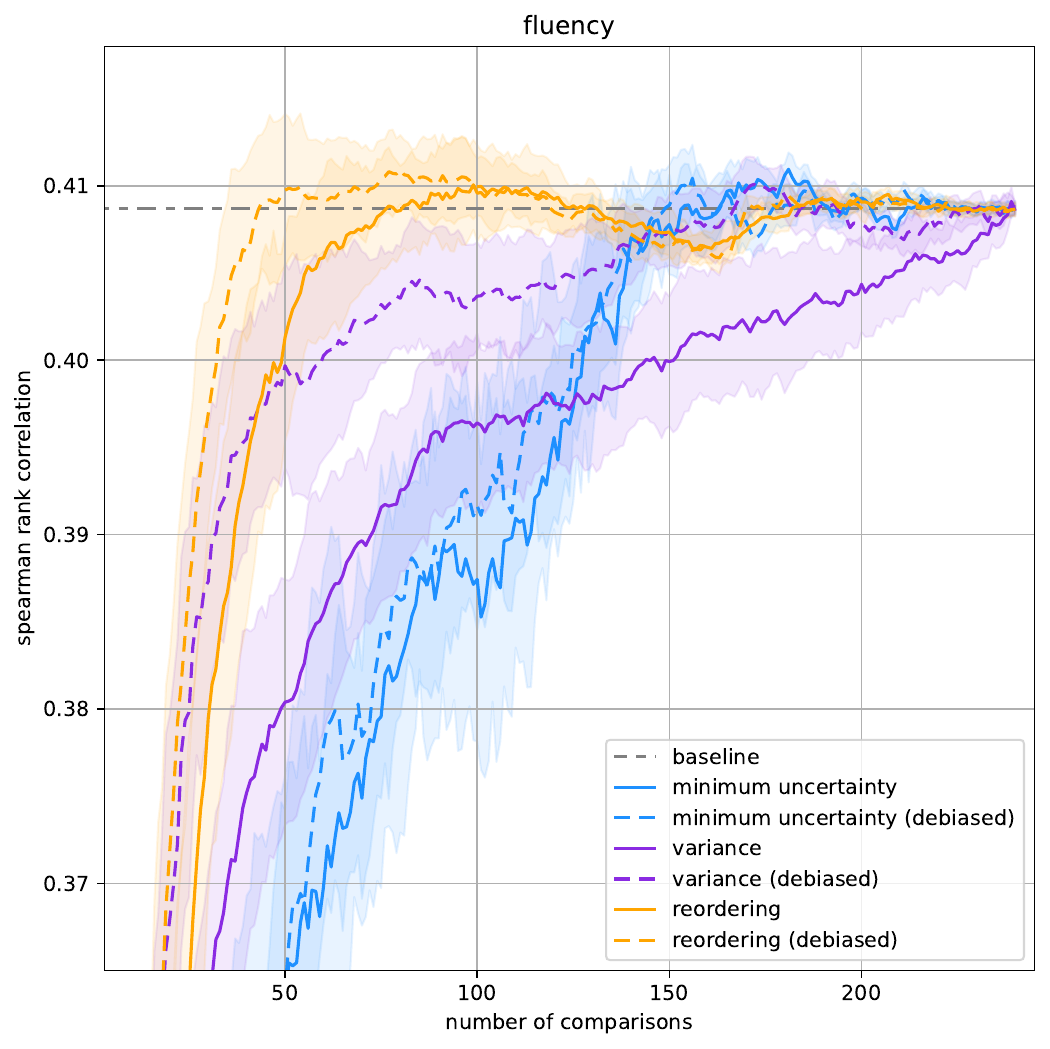}
        \label{fig:sel-flu}
    \end{subfigure}
    \hfill
    \begin{subfigure}{0.49\textwidth}
        \includegraphics[width=\textwidth]{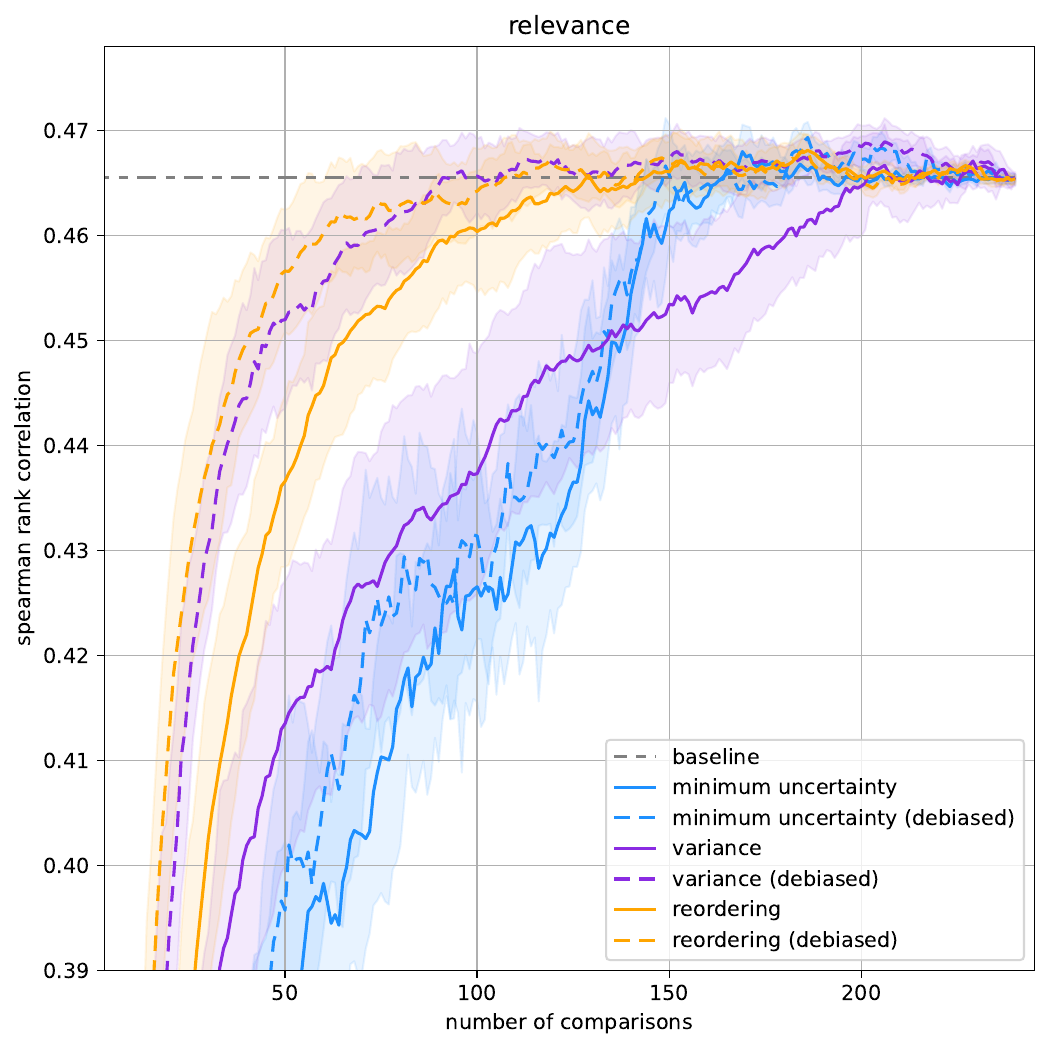}
        \label{fig:sel-rl}
    \end{subfigure}
    \vspace{-3mm}
    \caption{\textbf{Qwen2.5-3B-Instruct}: The Spearman Rank Correlation when iteratively selecting the next examples of lowest confidence/highest uncertainty. The baseline is the soft Bradley-Terry model with the minimum uncertainty metric. We also report the proposed variance and probability of reordering under the soft BT model.}
    \label{fig:selection-qwen-3b}
\end{figure*}

\begin{figure*}[h!]
    \begin{subfigure}{0.49\textwidth}
        \includegraphics[width=\textwidth]{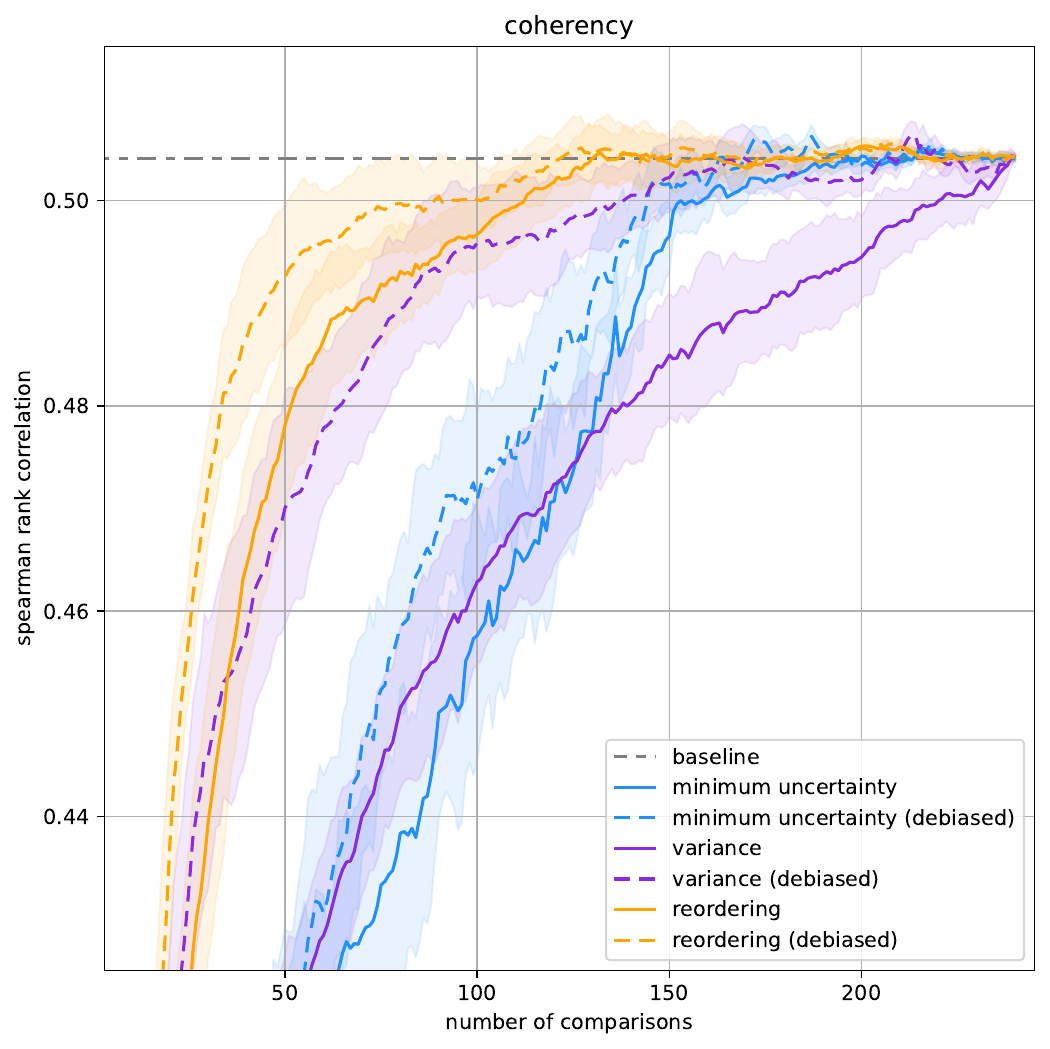}
        \label{fig:sel-coh}
    \end{subfigure}
    \hfill
    \begin{subfigure}{0.49\textwidth}
        \includegraphics[width=\textwidth]{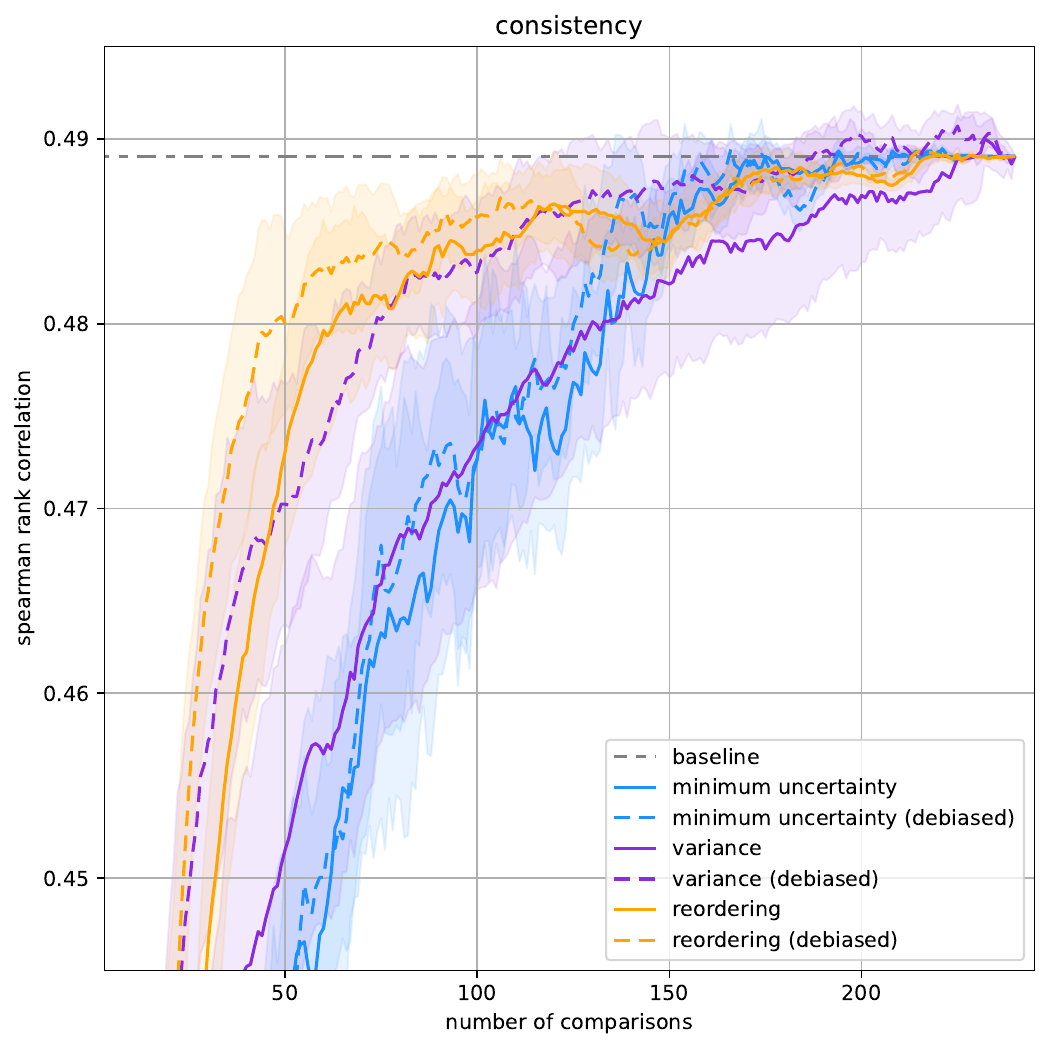}
        \label{fig:sel-con}
    \end{subfigure}
    \hfill
    \begin{subfigure}{0.49\textwidth}
        \includegraphics[width=\textwidth]{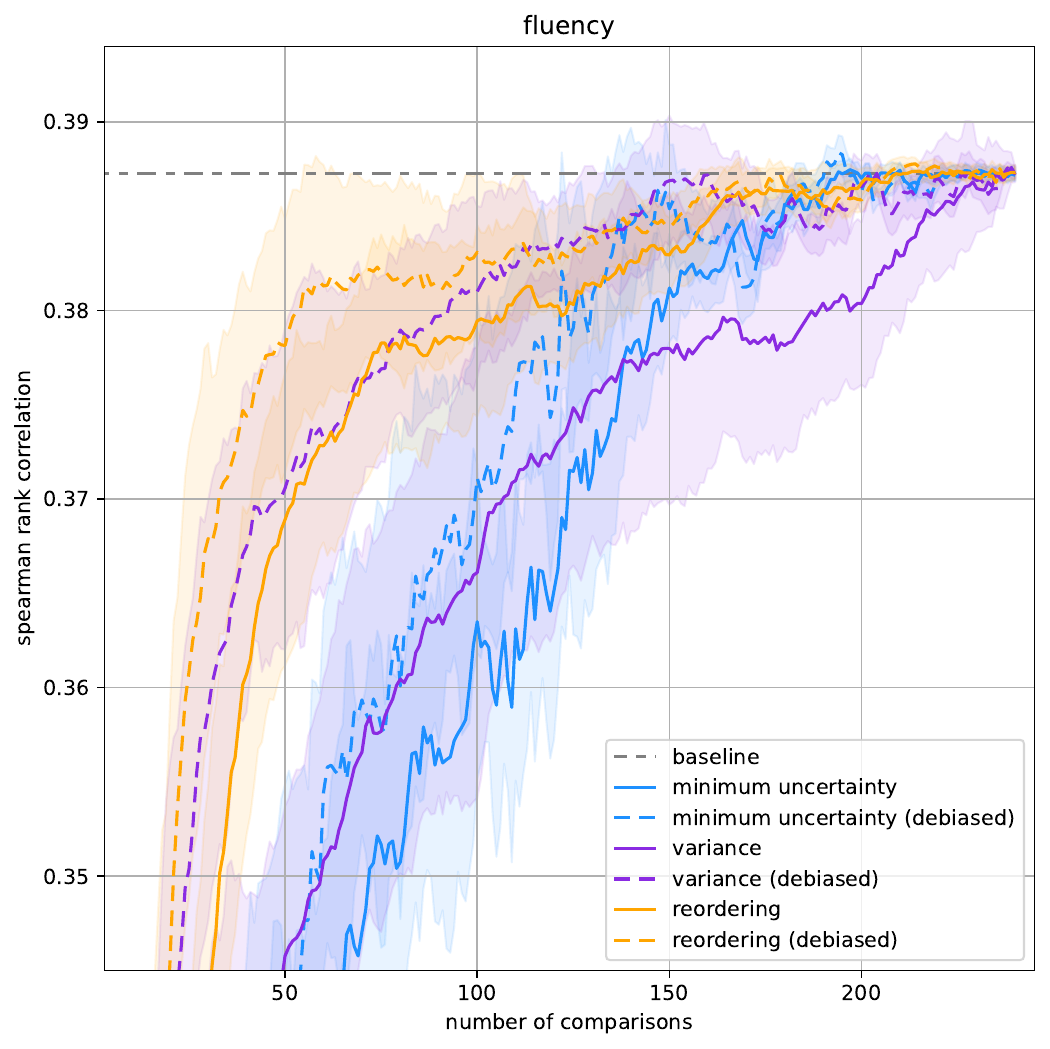}
        \label{fig:sel-flu}
    \end{subfigure}
    \hfill
    \begin{subfigure}{0.49\textwidth}
        \includegraphics[width=\textwidth]{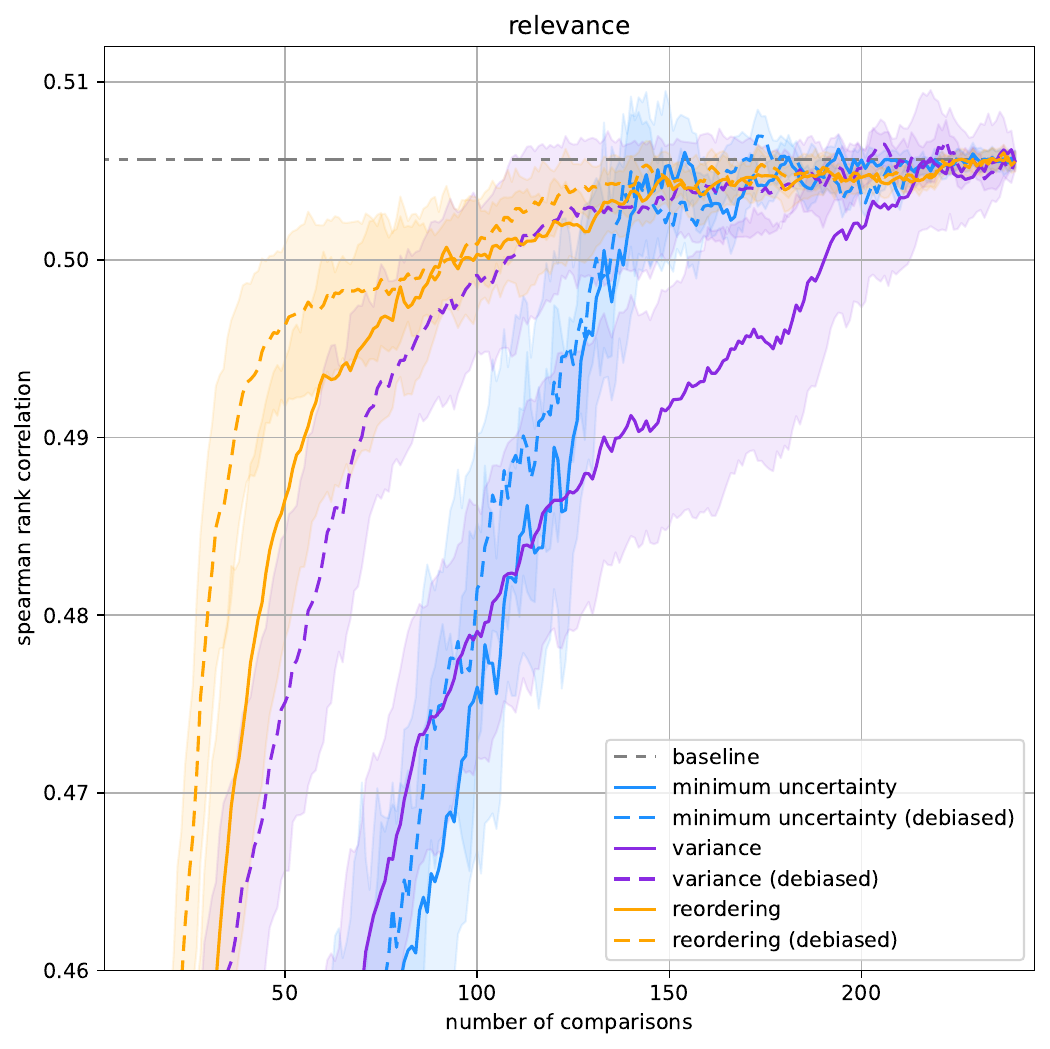}
        \label{fig:sel-rl}
    \end{subfigure}
    \vspace{-3mm}
    \caption{\textbf{Qwen2.5-7B-Instruct}: The Spearman Rank Correlation when iteratively selecting the next examples of lowest confidence/highest uncertainty. The baseline is the soft Bradley-Terry model with the minimum uncertainty metric. We also report the proposed variance and probability of reordering under the soft BT model.}
    \label{fig:selection-qwen-7b}
\end{figure*}

\begin{figure*}[h!]
    \begin{subfigure}{0.49\textwidth}
        \includegraphics[width=\textwidth]{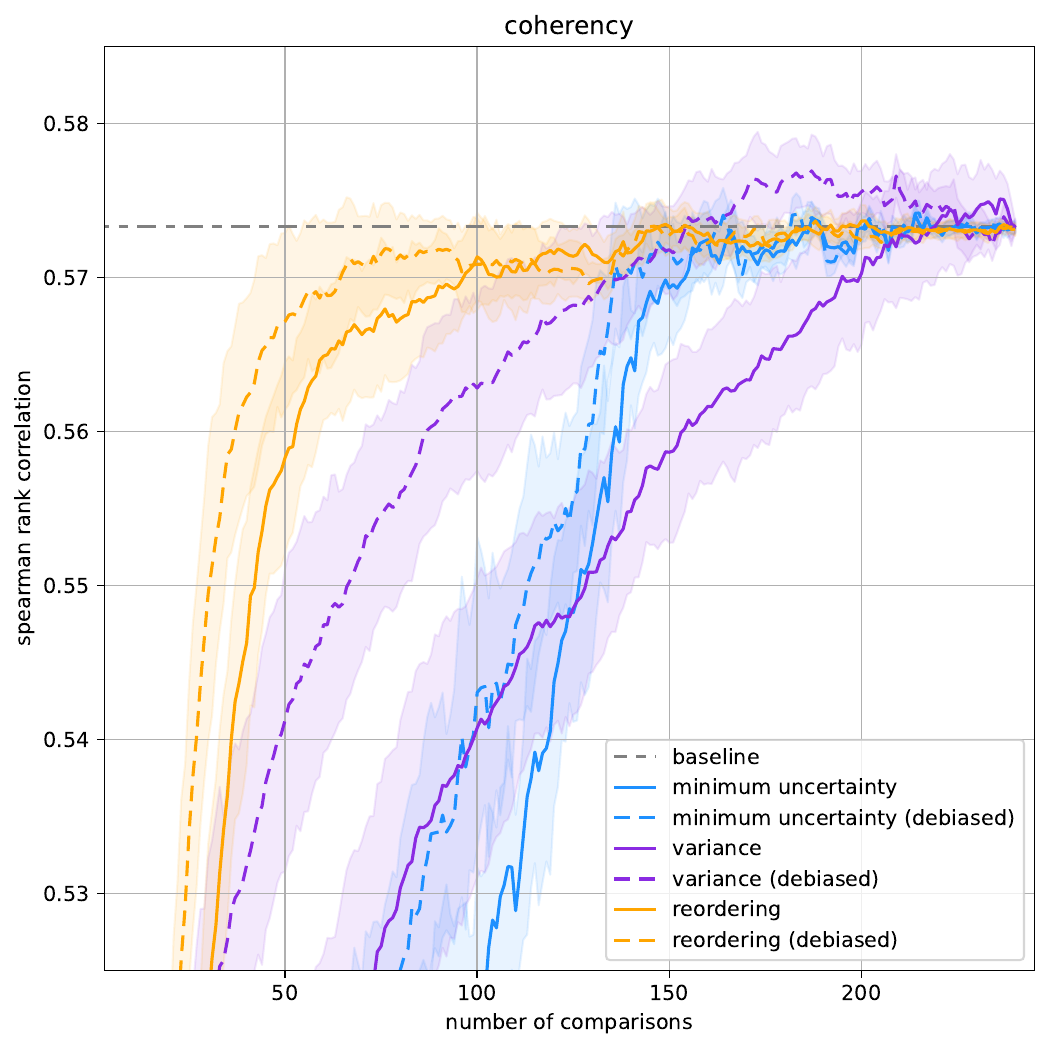}
        \label{fig:sel-coh}
    \end{subfigure}
    \hfill
    \begin{subfigure}{0.49\textwidth}
        \includegraphics[width=\textwidth]{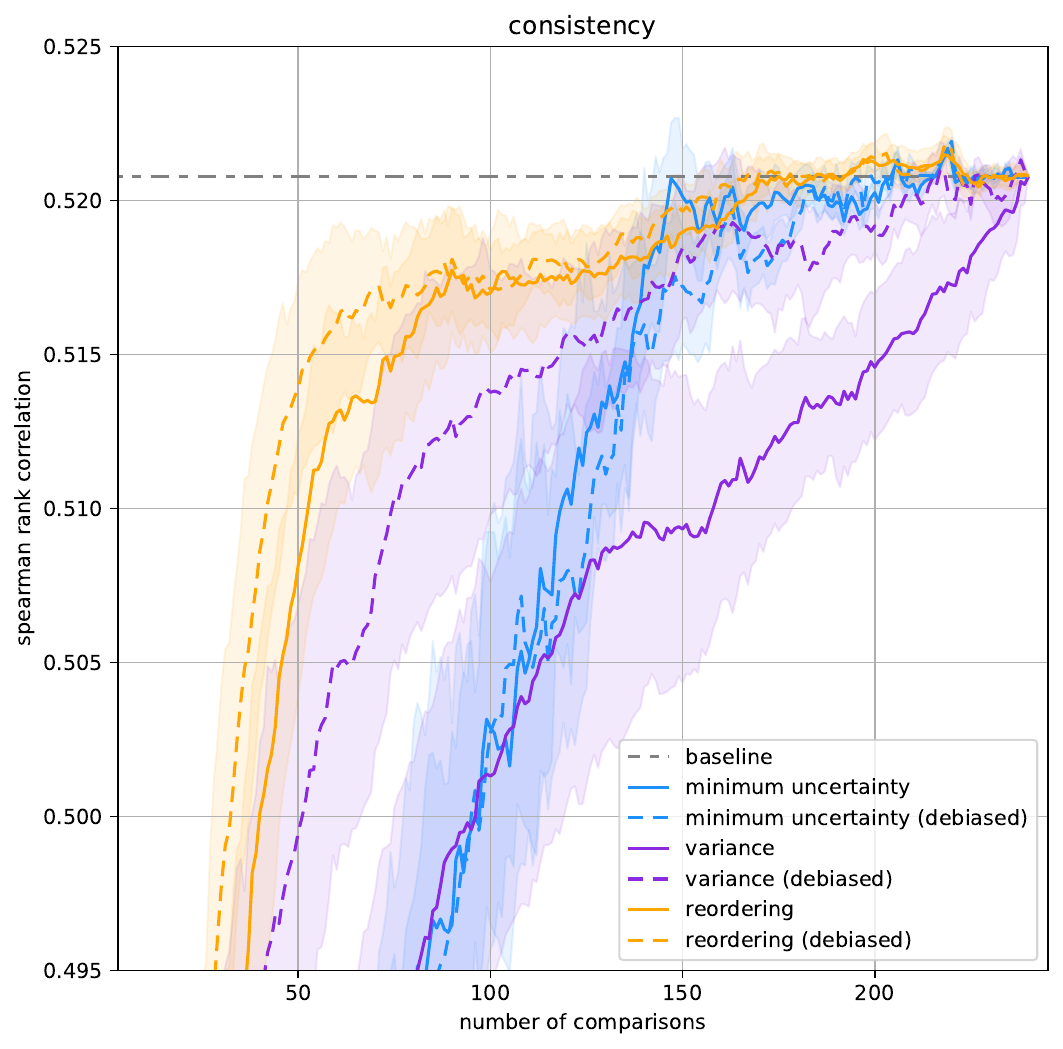}
        \label{fig:sel-con}
    \end{subfigure}
    \hfill
    \begin{subfigure}{0.49\textwidth}
        \includegraphics[width=\textwidth]{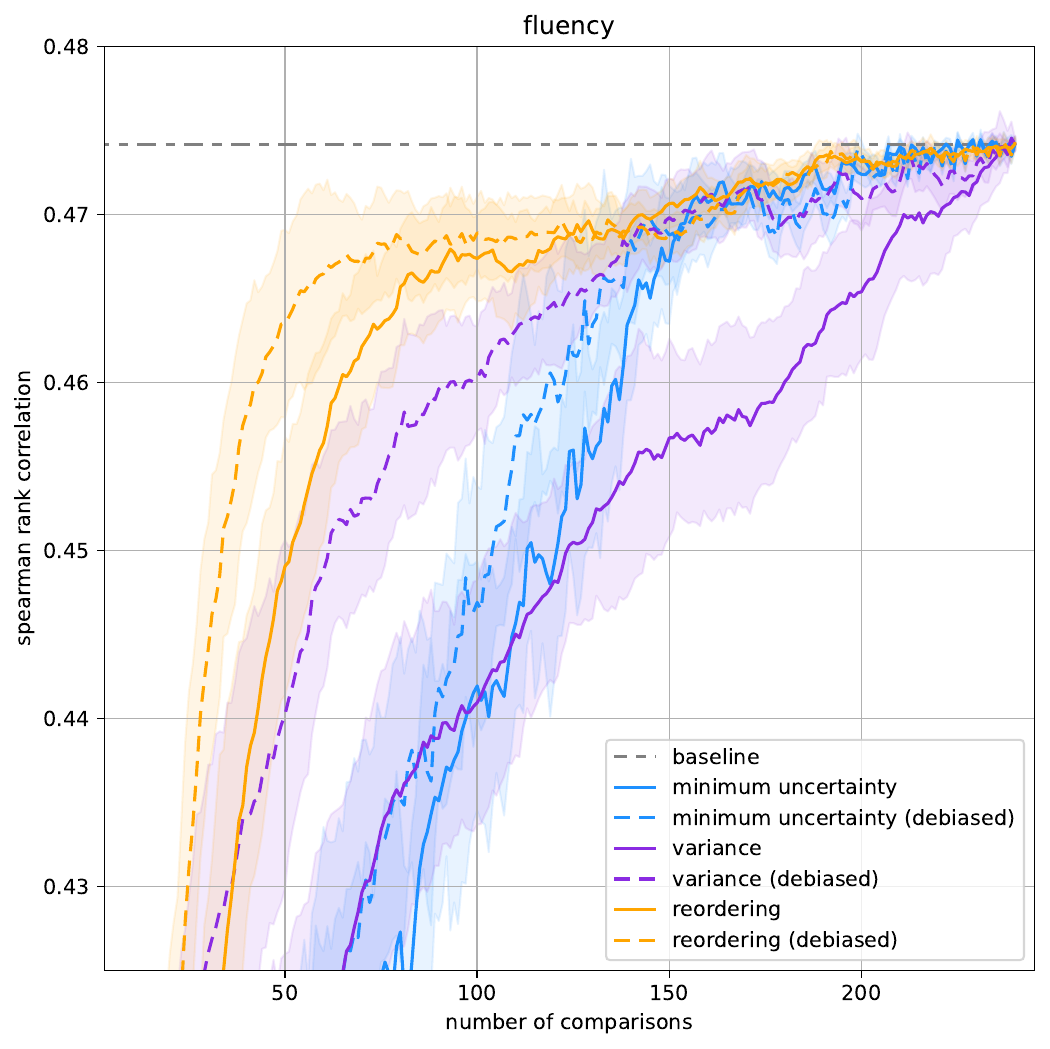}
        \label{fig:sel-flu}
    \end{subfigure}
    \hfill
    \begin{subfigure}{0.49\textwidth}
        \includegraphics[width=\textwidth]{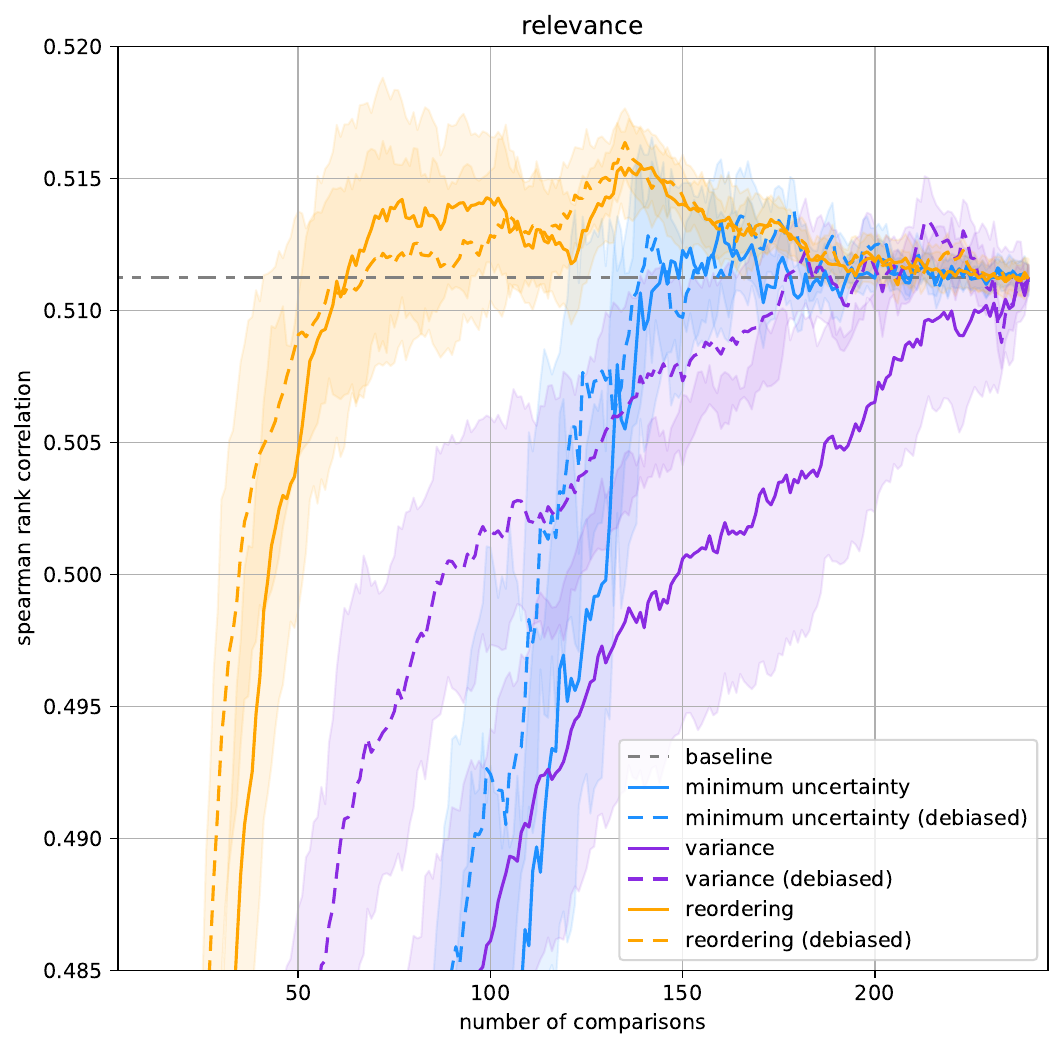}
        \label{fig:sel-rl}
    \end{subfigure}
    \vspace{-3mm}
    \caption{\textbf{Qwen2.5-14B-Instruct}: The Spearman Rank Correlation when iteratively selecting the next examples of lowest confidence/highest uncertainty. The baseline is the soft Bradley-Terry model with the minimum uncertainty metric. We also report the proposed variance and probability of reordering under the soft BT model.}
    \label{fig:selection-qwen-14b}
\end{figure*}

This section expands on results in Section \ref{ssec:iterative} for Qwen models. Evident from Figures \ref{fig:selection-qwen-3b}, \ref{fig:selection-qwen-7b} \& \ref{fig:selection-qwen-14b}, probability of reordering is the most robust metric for efficiently picking comparisons. While variance was a good choice under Flan-T5, the metric under biased probabilities suffers significantly, underperforming the minimum uncertainty metric in many cases.

\clearpage

\clearpage
The efficiency as reported in Table \ref{tab:poe-eff-qwen} corroborates this observation. While variance only performs well under debiased outputs from an LLM, the probability of reordering is stable under both cases, significantly outperforming the baseline minimum uncertainty in all cases.

\begin{table*}[h!]
    \centering
    \caption{\textbf{Measure of efficiency}: The number of comparisons needed to achieve 90\% performance of the baseline soft Bradley-Terry system for each corresponding judge. The baseline is evaluated using all $N(N-1) = 240$ comparisons.}
    \vspace{-3mm}
    \small
    \begin{tabular}{cc|cccc|cccc}
        \toprule
        \multirow{2}{*}{Uncertainties} & \multirow{2}{*}{Debiased} & \multicolumn{4}{c|}{Qwen2.5 (7B)} & \multicolumn{4}{c}{Qwen2.5 (14B)} \\
        & & {\tt COH} & {\tt CON} & {\tt FLU} & {\tt REL} & {\tt COH} & {\tt CON} & {\tt FLU} & {\tt REL} \\
        \midrule
        minimum & \xmark 
        & 73.1 \std 9.0
        & 46.9 \std 5.6 
        & 60.8 \std 13.3 
        & 71.8 \std 8.4 
        & 75.9 \std 7.7 
        & 50.2 \std 6.0 
        & 64.2 \std 12.7 
        & 67.6 \std 9.8 \\
        uncertainty & \cmark 
        & 61.0 \std 9.3 
        & 46.8 \std 8.3 
        & 50.9 \std 9.3 
        & 60.3 \std 11.1 
        & 67.0 \std 5.0 
        & 48.8 \std 7.4 
        & 62.0 \std 9.1 
        & 61.6 \std 9.3 \\
        \midrule
        \midrule
        \multirow{2}{*}{variance} & \xmark 
        & 56.1 \std 9.5 
        & 36.6 \std 9.6 
        & 52.0 \std 19.6 
        & 63.4 \std 10.7 
        & 59.7 \std 12.6 
        & 36.2 \std 8.0 
        & 61.0 \std 14.3 
        & 51.0 \std 9.6 \\
        & \cmark 
        & 24.3 \std 3.6 
        & 20.6 \std 4.2 
        & 23.2 \std 5.9 
        & 31.8 \std 5.9 
        & 27.6 \std 4.5 
        & 22.8 \std 2.4 
        & 31.6 \std 9.5 
        & 31.4 \std 6.0 \\
        \midrule
        probability & \xmark 
        & 35.6 \std 5.7 
        & 27.4 \std 2.7 
        & 31.6 \std 5.0 
        & 29.4 \std 4.3 
        & 28.1 \std 2.6 
        & 25.6 \std 2.5 
        & 34.2 \std 4.9 
        & 26.1 \std 2.7 \\
        of reordering & \cmark 
        & 21.5 \std 4.3 
        & 20.1 \std 2.5 
        & 20.5 \std 3.9 
        & 21.4 \std 2.4 
        & 20.6 \std 2.0 
        & 19.9 \std 2.1 
        & 23.1 \std 3.2
        & 20.1 \std 1.6 \\
        \bottomrule
    \end{tabular}
    \vspace{-2mm}
    \label{tab:poe-eff-qwen}
\end{table*}

\end{document}